\documentclass[lettersize,journal]{IEEEtran}
\usepackage{amsmath,amsfonts}
\usepackage{booktabs}
\usepackage{multirow}
\usepackage{multicol}
\usepackage{algorithmic}
\usepackage{algorithm}
\usepackage{array}
\usepackage[caption=false,font=normalsize,labelfont=sf,textfont=sf]{subfig}
\usepackage{textcomp}
\usepackage{stfloats}
\usepackage{url}
\usepackage[hidelinks]{hyperref}
\usepackage{verbatim}
\usepackage{graphicx}
\usepackage{cite}
\usepackage{arydshln}
\usepackage{bbding}
\usepackage[dvipsnames]{xcolor}
\usepackage{amssymb}
\newcommand{\ie}{\textit{i.e.,\ }} % TODO
\newcommand{\eg}{\textit{e.g.,\ }} % TODO

\begin{document}

\title{ROVER: Robust Loop Closure Verification with Trajectory Prior\\
in Repetitive Environments }
\author{
% anonymous authors
Jingwen Yu$^{1,2}$, \textit{Student Member, IEEE}, Jiayi Yang$^{3}$, Jianhao Jiao$^{5,6}$, \textit{Member, IEEE}, \\ Anjun Hu$^{4}$, Zhonghang Liu$^{6}$, Jiankun Wang$^{4}$, \textit{Senior Member, IEEE}, \\Ping Tan$^{1,*}$, \textit{Senior Member, IEEE}, and Hong Zhang$^{2,*}$, \textit{Life Fellow, IEEE}

\thanks{$^{1}$
        CKS Robotics Institute, Hong Kong University of Science and Technology, Hong Kong SAR, China (jyubt@connect.ust.hk)}
\thanks{$^{2}$
        Shenzhen Key Laboratory of Robotics and Computer Vision, Southern University of Science and Technology, Shenzhen, China}
\thanks{$^{3}$
        The University of Tokyo, Tokyo, Japan}
\thanks{$^{4}$
        Department of Electronic and Electrical Engineering, Southern University of Science and Technology, China}
\thanks{$^{5}$
        Department of Computer Science, University College London, UK}
\thanks{$^{6}$
        Department of Aeronautical and Aviation Engineering, The Hong Kong Polytechnic University, Hong Kong, China (jianhjiao@polyu.edu.hk)}
\thanks{${^7}$
        SmartMore Corporation Limited, Shenzhen, China}
\thanks{$^{*}$
        Corresponding authors: hzhang@sustech.edu.cn and pingtan@ust.hk
        }
\thanks{
This work was supported in part by Shenzhen Science and Technology Program (No. SGDX20240115111759002), in part by Meituan Academy of Robotics Shenzhen, in part by the Shenzhen Association for Science and Technology (No. XHXS2025-003), in part by National Natural Science Foundation of China under Grant (62473191, 62561160114), and in part by High level of special funds (G03034K003) from Southern University of Science and Technology, Shenzhen, China.
}
}

\markboth{Journal of \LaTeX\ Class Files,~Vol.~14, No.~8, August~2021}%
{Shell \MakeLowercase{\textit{et al.}}: A Sample Article Using IEEEtran.cls for IEEE Journals}


\maketitle

\begin{abstract}
Loop closure detection is important for simultaneous localization and mapping (SLAM), which associates current observations with historical keyframes, achieving drift correction and global relocalization. 
However, a falsely detected loop can be fatal, and this is especially difficult in repetitive environments where appearance-based features fail due to the high similarity.
Therefore, verifying a loop closure is a critical step to avoid false-positive detections.
Existing works in loop closure verification predominantly focus on learning invariant appearance features, neglecting the prior knowledge of the robot's spatial-temporal motion cue, i.e., trajectory.
In this article, we propose ROVER, a loop closure verification method that leverages the historical trajectory as a prior constraint to reject false loops in challenging repetitive environments.
For each loop candidate, it is first used to estimate the robot trajectory with pose-graph optimization. This trajectory is then submitted to a scoring scheme that assesses its compliance with the trajectory without the loop, which we refer to as the trajectory prior constraint (TPC), to determine if the loop candidate should be accepted.
Benchmark comparisons and real-world experiments demonstrate the effectiveness of the proposed method.
Furthermore, we integrate ROVER into state-of-the-art SLAM systems to verify its robustness and efficiency. 
Our source code and self-collected dataset will be made available online at https://rover-lcv.github.io/ upon publication of this article.
\end{abstract}
\def\abstractname{Note to Practitioners}
\begin{abstract}
This paper addresses a critical reliability challenge for autonomous mobile robots operating in repetitive environments, such as large warehouses, long corridors, or symmetrical structures. 
In these settings, high appearance similarity often causes robots to misidentify distinct locations. 
If the localization system incorrectly merges these locations (a ``false loop closure''), it can severely corrupt the robot's map and trajectory, leading to navigation failures that require manual intervention.
The primary contribution of this work is ROVER, a verification method that acts as a safeguard for simultaneous localization and mapping (SLAM) systems. 
Unlike traditional methods that rely primarily on appearance to validate loop closures, ROVER assesses the geometric consistency of the robot's historical motion. 
It effectively simulates the optimization that would occur if a match were accepted; if the resulting trajectory contradicts the robot's known motion history—implying physically impossible jumps or distortions—the match is rejected.
For practitioners, this approach offers a cost-effective method to enhance system robustness without requiring additional sensors or computationally expensive semantic understanding. 
While the system integration presented focuses on vision-based systems, the core methodology is sensor-agnostic and compatible with existing graph-based SLAM backends. 
It is important to note that ROVER relies on uncertainty estimates (covariance) for the robot's trajectory, which can be reliably estimated by practitioners for typical SLAM systems. 
The integration into proprietary navigation stacks should be straightforward with the provided source code to facilitate rapid prototyping and deployment.
\end{abstract}
\begin{figure}[!t]
    \centering
    \includegraphics[width=0.48\textwidth]{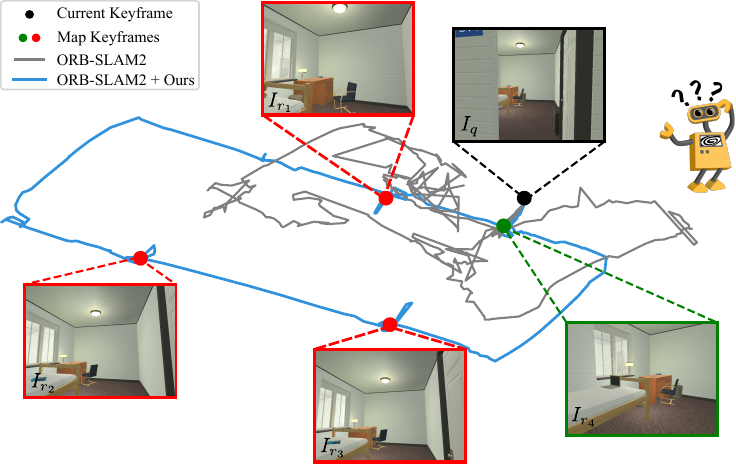}
    \caption{\textbf{A typical repetitive scenario} where a SLAM algorithm (e.g., ORB-SLAM2) fails for appearance-based loop verification. $I_{q}$ is the current keyframe (i.e., query image) used to detect loop closures, $I_{r_i}$ are the retrieved images among the map keyframes (i.e., reference database).
    Due to the visual similarity, geometric verification cannot disambiguate between
    true positives~\textcolor{OliveGreen}{\large\textbullet} and false positives~\textcolor{red}{\large\textbullet}.
    However, the proposed method incorporates trajectory cues to robustly reject
    false loops.}
    \label{fig:teaser}
\end{figure}
\section{Introduction}
\label{sec:intro} 
With SLAM advancing into the ``robust-perception age"~\cite{cadena2016past}, and ``lifelong navigation age"~\cite{yin2025general}, its deployment in unconstrained environments demands higher levels of robustness and reliability. Loop closure detection (LCD) has proven to be highly critical in ensuring global map consistency~\cite{campos2021orb,murai2024_mast3rslam}, achieving submap merging~\cite{campos2021orb,zou2024lta,wei2024large}, and recovery from tracking failure~\cite{chen2018submap, rahman2024large}.
LCD identifies the associations between a pair of previously visited and current
locations, subsequently incorporating the constraint into the SLAM back-end for
optimization. However, incorrect loop closure constraints can have fatal
consequences in robot state estimation, leading to its failure. The typical LCD
pipeline, which consists of a retrieval and a verification stage, is illustrated
in Fig.~\ref{fig:lcd}. In the retrieval stage, compact global descriptors are extracted
and compared to find the top-$K$ candidates for the verification stage, which
robustly rejects false candidates. While prior research has mainly focused on the
retrieval stage to enhance recall, the verification stage, that is crucial for ensuring
precision, remains largely underexplored. In this article, we focus on the
verification stage, specifically loop closure verification (LCV), a binary
classifier designed to robustly reject false loop closure candidates.

Currently, LCV is achieved mainly through geometric verification (GV), which has been exhaustively benchmarked in GV-Bench~\cite{yu2024gv}.
The existing challenges for GV that have been reported include repetitive patterns,
illumination changes, dynamic scene changes, and large viewpoint variations. Existing works focus on rejecting false loop closures via local features as described in Sec .~\ref{related:lcv}, such as local keypoints matching~\cite{yu2024gv}, objects matching~\cite{li2024resolving}, and text information~\cite{tao2025textinplace}. These methods predominantly rely on the similarity in appearance, neglecting the historical motion cue of the robot.
Appearance-based methods suffer from severe perceptual aliasing in repetitive
environments, as shown in Fig.~\ref{fig:teaser}. For LiDAR-based LCD,
verification mainly employs a simple fitness check using ICP registration~\cite{cadena2016past},
which also suffers from issues in repetitive environments~\cite{jin2024robust}.

Beyond loop closure verification, robust Pose Graph Optimization (PGO) methods
have been developed to mitigate the impact of false positive detections from LCD.
Common strategies include incorporating robust loss functions and solvers~\cite{Yang20ral-GNC}. Moreover, Vertigo~\cite{sunderhauf2012switchable} treats loop closure confidence as an optimizable latent variable. ~\cite{mangelson2018pairwise} models outlier rejection in multi-robot map merging as a maximum clique problem to extract the largest set of internally consistent measurements. iMCB-PGO~\cite{chen2024imcb} and Kimera-RPGO~\cite{Rosinol20icra-Kimera} adopt this concept for PGO to robustly reject spurious detections.
However, these methods are primarily designed for post-processing rather than online
incremental optimization. They are often sensitive to high outlier ratios and
require complex hyperparameter tuning, which can lead to failures in repetitive environments,
as benchmarked in Fig.~\ref{fig:backends}. The complex design of robust
optimization frameworks imposes additional computational costs. By leveraging LCV
to eliminate outliers prior to the backend, PGO can effectively refine large-scale,
dense graphs with a simple loss function and solver, achieving robustness and
efficiency.

We draw inspiration from Vertigo~\cite{sunderhauf2012switchable}, which observes
that a subset of false-positive loops within a given set of loop candidates can be
identified by analyzing the changes in residual error as the candidate loops are
selectively introduced to the PGO. In practical SLAM, however, loop candidates
are detected sequentially, necessitating that LCV considers one loop at a time.
In contrast, our work proposes to examine the changes in the robot's trajectory before
and after incorporating a loop constraint into PGO. Intuitively, adding a correct
loop to PGO, the change of the robot trajectory would be continuous and graceful;
while adding a false loop, the change would be chaotic and catastrophic~\cite{latif2013robust}.
By leveraging this prior knowledge in LCV, the challenge in repetitive
environments can be largely circumvented.

This paper aims to address the challenge of loop closure verification in
repetitive environments. We propose a novel method that is independent of sensor
measurements and leverages only the robot's historical trajectory. Since the
proposed method does not explicitly rely on sensor measurements, it is inherently
sensor-agnostic and can be integrated into any SLAM system, including visual
SLAM, LiDAR SLAM, Radar SLAM, etc. In this article, we focus on demonstrating the
effectiveness of the proposed method within the visual SLAM frameworks.

Our contributions can be summarized as follows:
\begin{enumerate}
    \item We propose a novel \textbf{RO}bust loop \textbf{VER}ification method (ROVER)
        that leverages the robot's historical trajectory rather than relying on
        appearance to robustly reject false positive loop detections.

    \item We propose to evaluate geometric consistency between the robot's trajectory
        before and after the PGO as the trajectory prior constraint (TPC), which
        serves as a measure of the confidence in loop candidates. This process is
        only related to the trajecotry, thus can be sensor-agnostic.

    \item We integrate ROVER tightly into SLAM systems to demonstrate the proposed
        method's robustness and efficiency through extensive experiments on public
        datasets and real-world experiments. The source code will be made
        publicly available for reimplementation and further research.
\end{enumerate}
The rest of the paper is organized as follows. Section~\ref{sec:related} reviews
the relevant literature on localization and mapping systems in repetitive environments,
loop closure detection, and loop closure verification methods. Section~\ref{sec:method}
overviews the proposed system and presents the core technical contribution of
ROVER. Section~\ref{sec:exp} demonstrates the experimental results, followed by
discussion and conclusion in Section~\ref{sec:discussion} and~\ref{sec:con}.
\section{Related Work}
\label{sec:related}
\subsection{Localization and Mapping in Repetitive Environments}
Existing visual SLAM systems typically depend on sparse local features~\cite{murORB2,xu2025airslam} or photometric consistency~\cite{huang2020monocular,lipson2024deep,teed2021droid}. Recently, VGGT-SLAM~\cite{maggio2025vggt-slam} and MASt3R-SLAM~\cite {murai2024_mast3rslam} leverage 3D foundation models to estimate dense correspondences for the front-end.
Existing SLAMs' loop closure detection (LCD) components struggle in environments with highly repetitive patterns, such as office buildings, warehouses, and underground parking lots, as shown in Fig.~\ref{fig:trajectory}.
However, LCD is critical for SLAM's global consistency. Thus, existing methods exploit extra sensors prior to providing global positioning assistance for SLAM in repetitive environments. Artificial landmarks~\cite{xu2022bifocal}, magnetic fields (MF)~\cite{wu2023global}, and radio frequency (RF) signals~\cite{zhang2021conquering} are employed as global reference to improve localization accuracy. Additionally, approaches such as~\cite{puligilla2020topological} assume Manhattan world structural priors for the topological mapping of warehouse scenes.
However, these methods rely heavily on environmental modifications, additional sensors, or prior knowledge of specific environments, which limits their generalizability and increases deployment costs. In contrast, we keep the original LCD unchanged by proposing ROVER, a system that rejects false positive loop detections and seamlessly integrates with existing SLAM frameworks (as illustrated in Fig.~\ref{fig:overview}) without assuming specific environmental constraints. Our method offers both flexibility and effectiveness for real-time autonomous robot navigation.
\begin{figure}[t]
\centering
\includegraphics[width=0.48\textwidth]{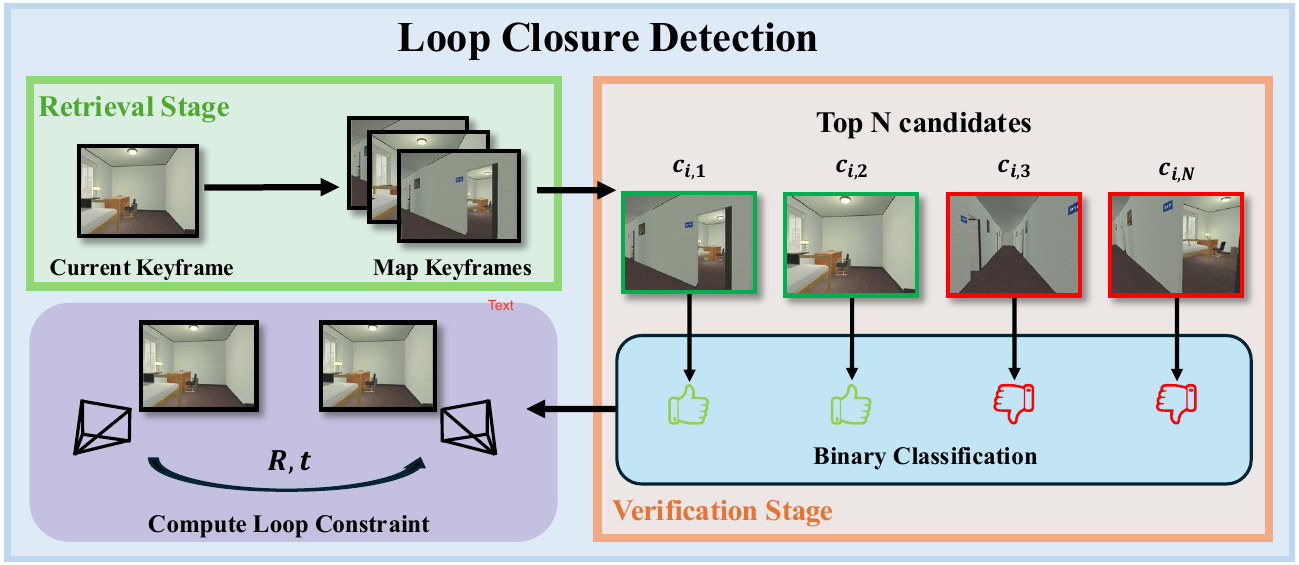}
\caption{\textbf{Two-stage Loop Closure Detection Pipeline.} i) Retrieval stage adopts image retrieval techniques to match historical keyframes. ii) The verification stage serves as a binary classifier that rejects the falsely detected loop candidates.}
\label{fig:lcd}
\end{figure}

\subsection{Loop Closure Detection}
\label{related:lcd}
Existing LCD methods in visual SLAM focus on appearance similarity, aiming to extract a global description of an image that is invariant to various conditions, e.g.,
viewpoint, illumination, season, and weather changes \cite{tsintotas2022revisiting}.
Traditionally, handcrafted features are used to achieve real-time LCD leading by DBoW2~\cite{angeli2008fast}. With the advent of deep learned representations, ~\cite{ye2023condition} takes advantage of visual latent encoding to achieve conditional invariance. With the invention of visual foundation models, AnyLoc~\cite{keetha2023anyloc} unleashes the large-scale pre-trained prior knowledge for place recognition, achieving outstanding performance for the retrieval stage of LCD.

As for LiDAR-based LCD, which has been extensively studied for large-scale SLAM for field robots. The scan-context (SC)~\cite{kim2018scan, kim2021scan} series is currently the most popular method, which projects the point cloud into a 2D plane and matches places as images.
While later works, such as STD~\cite{yuan2023std} and BTC~\cite{yuan2024btc}, propose encoding the 3D scene geometry in a compact global descriptor.  

These methods rely on rigid assumptions when evaluating appearance similarity for LCD, which often leads to failure in repetitive environments.
Thus, ~\cite{tao2025textinplace, jin2024robust} propose to leverage text information to disambiguate the repetitive loops. Specifically, an optical character recognition (OCR) is used to extract text entities and is fused with appearance-only descriptors. Their performance highly relies on the accuracy of OCR, which can easily fail when text entities do not exist or repeatedly appear (e.g., in the natural environments), as shown in Fig.~\ref{fig:handheld}(c) and Fig.~\ref{fig:pairwise}.

Despite the rapid development of these advanced methods, existing visual SLAM systems, such as VINS-Mono~\cite{qin2018vins}, ORB-SLAM2~\cite{murORB2}, and DPV-SLAM~\cite{lipson2024deep}, still rely on DBoW2 for real-time loop detection. To ensure robustness and efficiency, the proposed method focuses on the verification stage of loop detection. Our approach maintains the recall while increasing the precision by keeping the retrieval stage unchanged.

\begin{figure}[t!]
\centering
\includegraphics[width=0.48\textwidth]{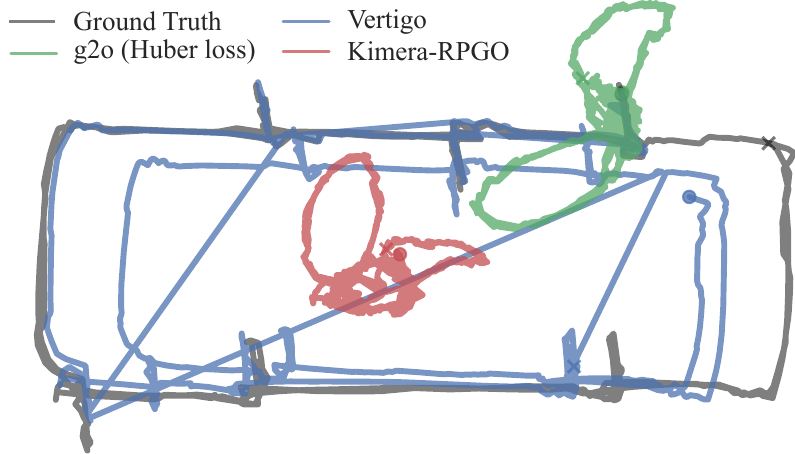}
\caption{\textbf{Comparison of different robust PGO backends} on Hotel~\cite{li2024resolving} dataset. The circles and crosses for each trajectory represent the start and end positions. The input pose graph is constructed using odometry estimation from ORB-SLAM2~\cite{murORB2} with loop constraints detected by DBoW2 with post-filtering by geometric verification using ORB features. Due to the high ratio of false-positive loop detections, robust backends fail to recover the trajectory through offline post-processing.}
\label{fig:backends}
\end{figure}
\subsection{Loop Closure Verification}
\label{related:lcv}
As shown in Fig.~\ref{fig:lcd}, the two-stage loop closure detection consists of a retrieval stage and a verification stage. Existing LCD methods primarily focus on the retrieval stage, either ignoring the verification stage or employing a standard method for verification, which is responsible for rejecting falsely retrieved candidates. In this section, we review the existing solutions designed for LCV.
Currently, GV, based on multi-view geometry, serves as the dominant solution to reject falsely detected loop candidates. 
Specifically, the epipolar constraint in two-view geometry is utilized to conduct RANSAC, and the number of inliers in local feature matches serves as an indication of the positive likelihood. 
GV has recently been greatly enhanced by learning-based local feature techniques~\cite{sarlin2020superglue, sun2021loftr, leroy2024grounding}. However, it still faces challenges in visually similar scenarios since the epipolar constraint fails due to repetitive patterns~\cite{yu2024gv}.
\cite{li2024resolving} targets rejecting the falsely detected loops with the assistance of a large language model (LLM). 
It first extracts the scene semantics (e.g., door numbers) and then uses a visual question and answer (VQA) model to generate a text description. Then it feeds the text description and tailored prompt to ChatGPT.
It faces challenges caused by repeated semantic entities and scenes without textual information, as shown in Fig.~\ref{fig:viz}(d).
Moreover, it heavily relies on the reasoning ability of an LLM, which is not certifiable and struggles to achieve real-time performance.

As an alternative to addressing the challenges of LCV in repetitive environments, we propose a novel approach that utilizes information from the robot's trajectory before and after PGO, rather than relying solely on appearance information within the current view and the loop candidate. As a result, it can effectively avoid ambiguous scenarios in repetitive scenes while offering seamless integration with existing SLAM systems.
\begin{figure*}[t]
\centering
\includegraphics[width=0.98\textwidth]{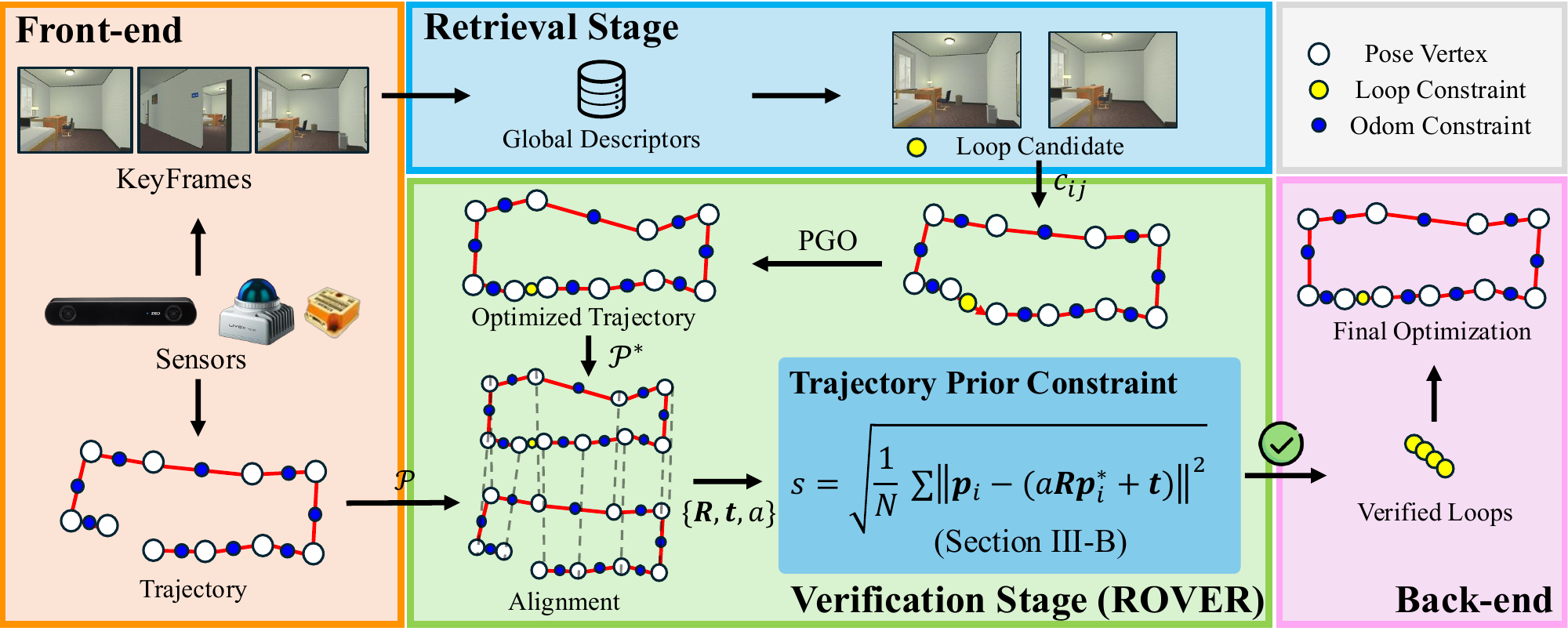}
\caption{\textbf{System overview} of a complete SLAM pipeline with proposed loop verification method. Our main contribution lies in the verification stage (as explained in Section~\ref{method:tpc}), which takes loop constraints from the retrieval stage and the estimated trajectory from the front-end to predict a confidence score for rejecting false loops. In the end, the verified loops are accepted for the final back-end optimization.} 
\label{fig:overview}
\end{figure*}

\section{Methodology }
\label{sec:method}
This section presents our proposed ROVER by first defining the notations used. Section~\ref{method:overview} overviews the system design as demonstrated in Fig.~\ref{fig:overview}, followed by a detailed description of the trajectory prior constraint (TPC) in Section~\ref{method:tpc}.
We further extend our work by integrating it into visual SLAM systems to validate its efficacy in Section~\ref{method:integration}. 
\subsection{Notations}
\label{sec:notation}
Throughout the paper, we denote the keyframe collected at the $k$-th time as $\mathcal{F}_k$.
A rigid transform $\mathbf{x}_k = [\mathbf{R}_k| \mathbf{p}_k] \in SE(3)$ represents the pose of $\mathcal{F}_k$ relative to $\mathcal{F}_{0}$. 
A robot trajectory can be represented by a set of keyframe poses denoted as $\mathcal{X} = \{\mathbf{x}_i\}_n$.
The translation part of the trajectory is denoted as $\mathcal{P} = \{\mathbf{p}_k\ | \mathbf{p}_k \in \mathbb{R}^3 \}$.
The pose graph $\mathcal{G} = (\mathcal{X}, \mathcal{U})$ consists of a set of nodes $\mathcal{X}$ and a set of noisy relative measurements $\mathcal{U} = \{ \mathbf{u}_{ij} | \mathbf{u}_{ij} \in SE(3) \}$ between $\mathcal{F}_i$ and $\mathcal{F}_j$.
\subsection{System Overview}
\label{method:overview}
The proposed pipeline is illustrated in Fig.~\ref{fig:overview}, where ROVER serves as an intermediate layer between front-end estimation and back-end optimization.
The proposed design follows the convention of the classical pose graph SLAM framework.
Thus, it can be integrated into the SLAM system regardless of sensing modality.
In this letter, we focus on the visual SLAM systems. 
The front-end odometry estimates the pose of keyframes $\{\mathcal{F}_i\}_n$, which produces $\mathcal{X}_n$ and corresponding odometry constraints $\{c_{i, i+1}\}_{n-1}$ from relative measurements $\mathcal{U}_n$:
\begin{equation}
    c_{i, i+1} = ||f(\mathbf{x}_i, \mathbf{u}_{i}) - \mathbf{x}_{i+1}||^{2}_{\boldsymbol{\Sigma_{i}}}.
\end{equation}
The loop detection module detects the top $k$ candidates for each keyframe, producing $k$ loop candidates $\{\mathcal{F}_i, \mathcal{F}_j \}_k$. For each candidate, a relative pose $\mathbf{u}_{ij}$ is estimated. Accordingly, a loop constraint $c_{ij}$ is generated:
\begin{equation}
    c_{ij} = ||f(\mathbf{x}_i, \mathbf{u}_{ij}) - \mathbf{x}_{j}||^{2}_{\boldsymbol{\Lambda_{ij}}}.
\end{equation}
The classical PGO is formulated as follows:
\begin{equation}
\label{equ:pgo}
\begin{split}
     \mathcal{X}^* = \operatorname*{argmin}_{\mathcal{X}} \underbrace{\sum_{i}^{n} ||f(\mathbf{x}_i, \mathbf{u}_{i}) - \mathbf{x}_{i+1}||^{2}_{\boldsymbol{\Sigma_{i}}}}_{\text{Odometry Constraints}}\\
    + \quad \underbrace{\sum_{ij}^{k} || (f(\mathbf{x}_i, \mathbf{u}_{ij}) - \mathbf{x}_{j})||^{2}_{\boldsymbol{\Lambda_{ij}}}}_{\text{Loop Constraints}}.
\end{split}
\end{equation}
ROVER takes $\mathcal{X}$ and loop constraint $c_{ij}$ as input, and predicts a loop verification score $s_{ij}$ based on the trajectory prior constraint (TPC) described in Section~\ref{method:tpc}, which can be injected into the PGO formulation seamlessly:
\begin{equation}
\label{equ:rover}
\begin{split}
     \mathcal{X}^* = \operatorname*{argmin}_{\mathcal{X}} \sum_{i}^{n} ||f(\mathbf{x}_i, \mathbf{u}_{i}) - \mathbf{x}_{i+1}||^{2}_{\boldsymbol{\Sigma_{i}}} \\
    + \quad || \Psi (s_{ij}) * (f(\mathbf{x}_i, \mathbf{u}_{ij}) - \mathbf{x}_{j})||^{2}_{\boldsymbol{\Lambda_{ij}}}.
\end{split}
\end{equation}
Note that our proposed system verifies the detected loop individually in a sequential manner as described in Equation~(\ref{equ:rover}). Loops are verified by thresholding the probability $\Psi:\mathbb{R} \to \{0, 1\}$. If a loop closure candidate fails to pass the threshold, it is rejected for further back-end optimization.
In this way, ROVER bridges the front-end odometry estimation and back-end optimization, building a robust SLAM framework.
\begin{figure}[ht]
\centering
\includegraphics[width=0.48\textwidth]{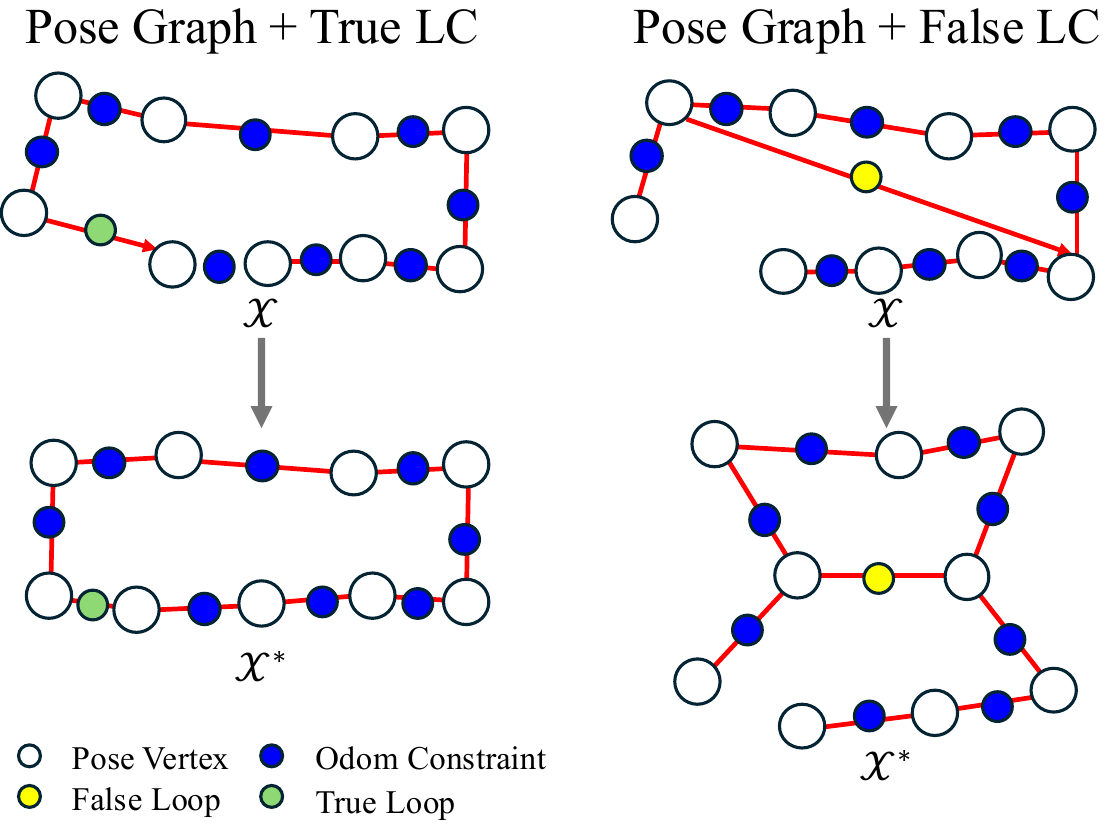}
\caption{\textbf{Illustrative example of TPC.} Adding a true loop to PGO would result in a continuous and graceful change, as shown on the left-hand side, while a false loop is more likely to result in a chaotic change, as shown on the right.}
\label{fig:method}
\end{figure}
\subsection{Trajectory Prior Constraint}
\label{method:tpc}
\begin{figure*}[ht]
\centering
\includegraphics[width=0.96\textwidth]{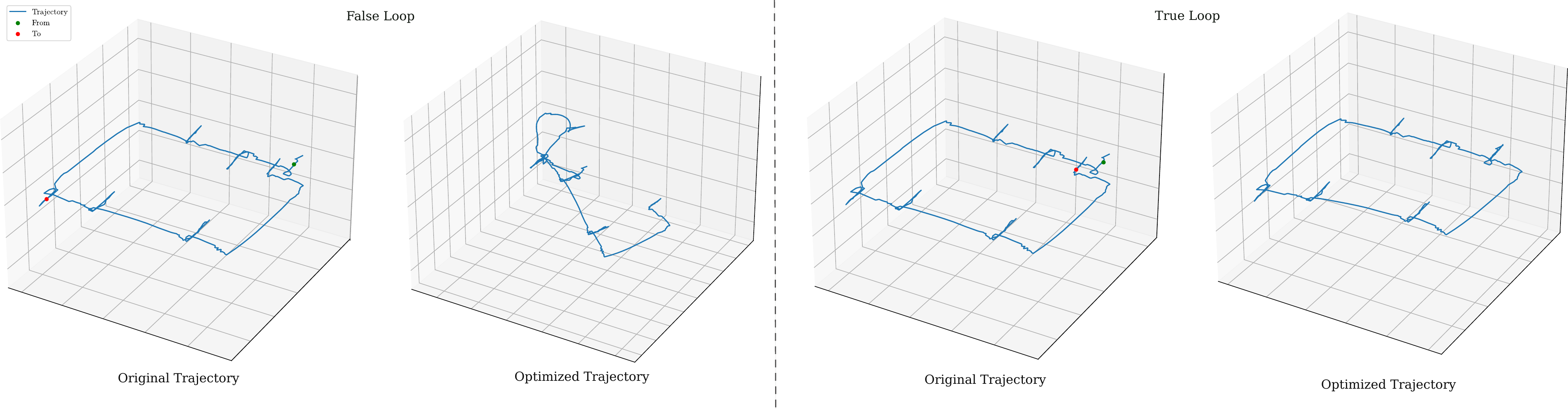}
\caption{\textbf{Visualization of true and false loop closures on the Hotel dataset.} To complement Fig.~\ref{fig:method}, we demonstrate two real-world trajectory pairs to illustrate the effectiveness of TPC. While a true loop introduces a geometrically ``elegant'' change between the original trajectory $\mathcal{X}$ and the optimized trajectory $\mathcal{X}^*$, a false loop instead causes severe, ``chaotic'' deformation.}
\label{fig:hotel_trajectory}
\end{figure*}

The trajectory inherently contains historical motion cues, which can serve as a strong prior for loop verification to enhance the precision of LCD. 
As discussed earlier in Section~\ref{sec:intro}, the key intuition is that a valid loop closure should correct drift, resulting in a graceful change rather than distorting the entire trajectory into a chaotic shape, as illustrated in Fig.~\ref{fig:method}. 
To this end, we propose measuring the change of the trajectory before and after PGO, i.e., between $\mathcal{X}$ and $\mathcal{X}^*$, under the assumption that the loop constraint $c_{ij}$ holds true.
The proposed scoring scheme is in two folds: i) align $\mathcal{X}$ and $\mathcal{X}^*$ based on the coherent timestamps, and ii) compute the change score of two aligned trajectories.
For the alignment of two trajectories, using only the translational components is sufficient, as discussed in~\cite{salas2015trajectory}. 
Therefore, we take $\mathcal{P}$ and $\mathcal{P}^*$, which represents the translational part of trajectories of $\mathcal{X}$ and $\mathcal{X}^*$, respectively.
We adopt~\cite{umeyama1991least} for alignment by finding the rigid $SIM(3)$ transform $\mathcal{S}^* = \{\mathbf{R}, \mathbf{t}, a\}$ that satisfies:
\begin{equation}
\label{equ:umeyama}
\mathcal{S}^* =\operatorname*{argmin}_{\mathcal{S}} \sum_{i=0}^{N-1} \left\| \mathbf{p}_i - a\mathbf{R}\mathbf{p}^*_i - \mathbf{t} \right\|^2.
\end{equation}
Then, the Euclidean distance between the aligned trajectories is taken as the confidence $s_{ij}$ for verification:
 \begin{equation}
\label{equ:score}
  s_{ij} =  \sqrt{\frac{1}{N} \sum_{i=0}^{N-1} \|\mathbf{p}_i - (a\mathbf{R}\mathbf{p}^*_i + \mathbf{t})\|^2},
\end{equation}
where $\{\mathbf{R}, \mathbf{t}, a \} \in \mathcal{S}^*$.
\begin{algorithm}[t!]
\caption{Loop Verification based on TPC}
\label{alg:tpc}
\begin{algorithmic}[1]
\STATE \textbf{Notation:} Pose Graph $\mathcal{G}$, Loop Candidate $c_{ij}$
\STATE \textbf{Input:} $\mathcal{X}$ and $c_{ij}$ 
  \STATE \textbf{Output:} Predicted label $\{0, 1\}$ for $c_{ij}$
  \STATE Initialize $\mathcal{G} \gets$ Front-end
  \STATE $\mathcal{X}^* \gets PGO(\mathcal{G}, c_{ij})$ \quad \textcolor{gray}{// Equation~(\ref{equ:pgo})}
  \IF{$PGO$ Converge} 
  \STATE $\{\mathcal{P}, \mathcal{P}^*\} \gets \{\mathcal{X}, \mathcal{X}^*\}$
  \STATE $\mathcal{S}^* = TrajAlign(\mathcal{P}, \mathcal{P}^*)$ \quad  \textcolor{gray}{// Equation~(\ref{equ:umeyama})}
  \STATE $s_{ij} = ComputeScore(\mathcal{P}, \mathcal{P}^*, \mathcal{S}^*)$ \quad \textcolor{gray}{// Equation~(\ref{equ:score})}
  \IF{$\Psi (s_{ij}) == 1$}
  \RETURN 1
  \ENDIF
\ENDIF
\RETURN 0
\end{algorithmic}
\end{algorithm}
The complete algorithm pipeline can be summarized in Algorithm~\ref{alg:tpc}. A pose graph $\mathcal{G}$ can be constructed from the front-end of SLAM following Equation~(\ref{equ:pgo}) along with a loop candidate constraint $c_{ij}$. 
The thresholding strategy of $\Phi(s_{ij})$ is designed heuristically by considering both the scene scale and uncertainty of the odometry estimator.
Notably, the proposed method is designed for sequential input, i.e., loop candidates are verified one by one, a scenario where PGO methods that require batch input typically fail.
Moreover, accurate verification results can improve the PGO's scalability and efficiency by pre-filtering outliers.
Finally, the verification is accomplished by first aligning the translational part of the trajectories $\mathcal{P}$ and $\mathcal{P}^*$ using Equation~(\ref{equ:umeyama}) and predicting the score using Equation~(\ref{equ:score}).
In Fig.~\ref{fig:hotel_trajectory}, real examples of the trajectory deformation are demonstrated as the complement to the illustration of Fig.~\ref{fig:method}. The false loop causes significant deformation to the original trajectory, where the true loop introduces a geometrically consistent change between $\mathcal{X}$ and $\mathcal{X}^*$.

\subsection{Integration into SLAM Systems}
\label{method:integration}
LCD is closely tied to SLAM systems, as variations in keyframe selection strategies can significantly impact loop detection results. To highlight the effectiveness of ROVER, we evaluate it alongside real-time SLAM systems. ORB-SLAM2 and VINS-Fusion are selected as baselines for experiments, representing state-of-the-art (SOTA) visual SLAM and visual-inertial SLAM approaches, respectively.
While the proposed method is in general compatible with any sensor modality, this section focuses on typical visual SLAM systems as examples. 
ORB-SLAM2 represents the keyframe-based visual SLAM approach, featuring a dedicated \textit{LoopClosing} thread for loop retrieval, verification, and correction.
Similarly, VINS-Fusion includes a \textit{LoopFusion} thread that functions in a manner analogous to ORB-SLAM2's \textit{LoopClosing} thread.
Here, we use the \textit{LoopClosing} thread as an example to demonstrate the integration of ROVER into the SLAM pipeline in Algorithm~\ref{alg:loopclosing}.
\begin{algorithm}[t!]
\caption{\textit{LoopClosing} Thread with ROVER}
\begin{algorithmic}[1]
\label{alg:loopclosing}
\STATE \textbf{Notation:} Map Keyframes $\mathcal{K}=\{\mathcal{F}_k\}$, Latest Keyframe $\mathcal{F}_i$
\STATE Initialize: $\mathcal{F}_i \gets$ \textit{Tracking} Thread
  \WHILE{$\mathcal{K} \neq \varnothing $}
  \IF{$CheckNewKeyFrame(\mathcal{F}_i)$}
  \STATE  $\mathcal{F}_j \gets DetectLoop(\mathcal{F}_i)$
  \IF{$ 0 \leq j < i $ }  
  \STATE $c_{ij}, n_{matches} \gets ComputeSim3(\mathcal{F}_i, \mathcal{F}_j)$
  \IF{$n_{matches} \geq 40$} 
  \STATE $s_{ij} \gets ROVER(\mathcal{G}, c_{ij})$
  \IF{$\Psi (s_{ij}) == 1$}
  \STATE $CorrectLoop()$ 
  \ENDIF
  \ENDIF
  \ENDIF
  \ENDIF
  \ENDWHILE
\end{algorithmic}
\end{algorithm}
When a new valid keyframe $\mathcal{F}_i$ arrives, DBoW2 is employed to retrieve the closest historical keyframe $\mathcal{F}_j$. Then, \textit{ComputeSim3} returns a loop constraint $c_{ij}$ by estimating the relative pose from local feature matching, if $\mathcal{F}_i$ and $\mathcal{F}_j$ pass the GV. 
In default ORB-SLAM2, the threshold for GV is set to 40 inliers. 
The loop is further verified by ROVER given the pose graph $\mathcal{G}$ and loop constraint $c_{ij}$.
Finally, the verified loop will be used for loop correction. 
The $LoopFusion$ thread of VINS-Fusion follows a similar logic as described above in Algorithm~\ref{alg:loopclosing}.
Note that for SLAM systems where the absolute scale is observable (\eg, VIO and stereo visual SLAM), $\mathcal{S}$ in Equation (\ref{equ:umeyama}) reduces to $SE(3)$, \ie $a=1$.
For implementation details, we refer readers to our open-source code.
\section{Experimental Results}
\label{sec:exp}
\begin{figure*}[t!]
    \centering
    \includegraphics[width=0.98\textwidth]{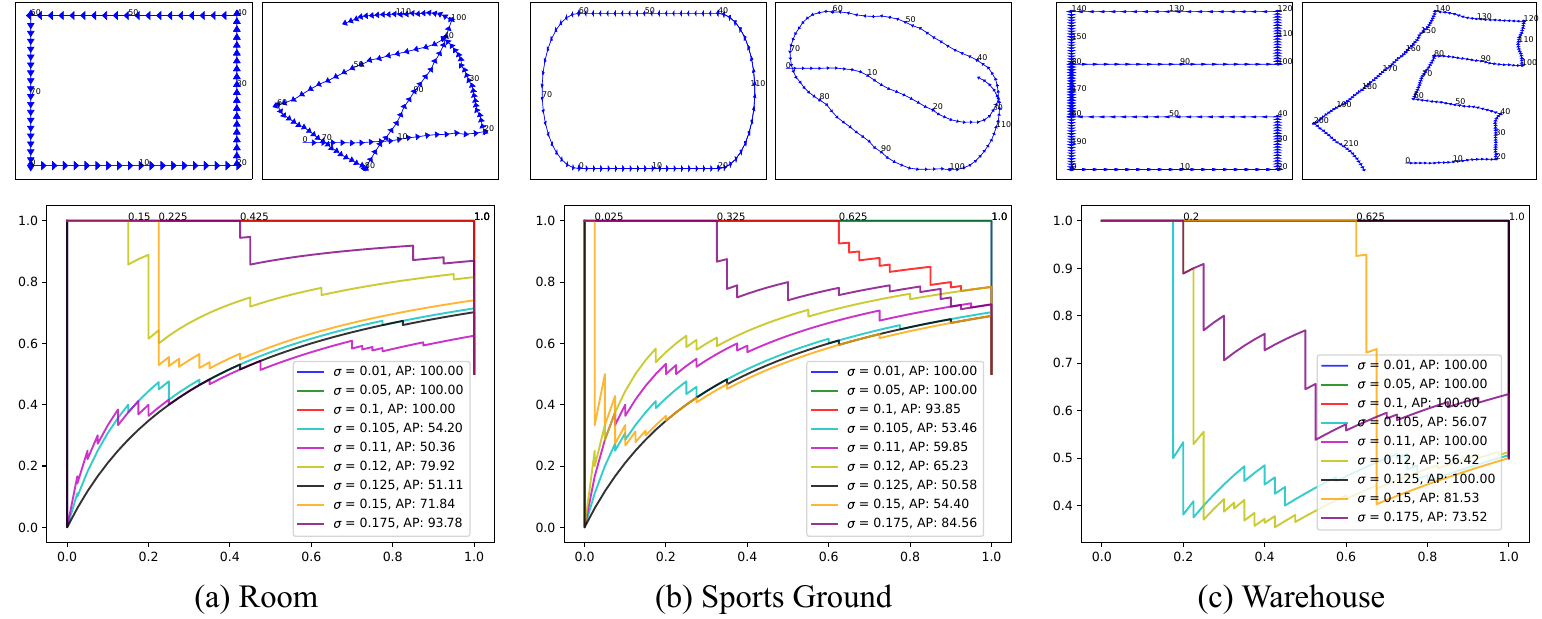}
    \caption{\textbf{Proof-of-Concept Experiment.} We have simulated three common scenarios: (a) Room, (b) Sports Ground, and (c) Warehouse. For each scenario, we generate a ground truth trajectory (upper left) and simulated odometry with a Gaussian noise level $\sigma=0.1$ (upper right), and Precision-Recall curves of the proposed method under different odometry noise levels $\sigma\in[0.01, 0.175]$ (bottom).
    The true positive (TP) and false positive (FP) verification candidates are generated randomly as the input. The proposed method consistently achieves high recall and precision across different noise levels, demonstrating robustness across different trajectory shapes.}
    \label{fig:poc}
\end{figure*}
In this section, we first introduce a set of proof-of-concept experiments in Section~\ref{sec:poc} to demonstrate
the rationale of the foundation design of our approach in Fig.~\ref{fig:poc}.
Datasets and metrics used for experiments are described in Section~\ref{exp:datasets},
including a self-collected one with a customized hardware design. Subsequently,
we conduct comprehensive experiments from two perspectives:
\begin{enumerate}
    \item Loop closure verification (Section~\ref{exp:LCV}): We compare the
        proposed work with GV baselines, which are normally employed as the verification
        stage of LCD. Moreover, SOTA global retrieval methods are benchmarked in
        reference to the LCD's retrieval stage.

    \item Localization accuracy (Section~\ref{exp:SLAM}): We evaluate the
        localization accuracy by integrating the proposed method with visual
        SLAM baselines to demonstrate the practical efficacy.
\end{enumerate}
Finally, we conclude this section with an efficiency evaluation in Section~\ref{sec:eff}, as the importance of real-time performance in SLAM systems is emphasized. Existing limitations and potential directions for future improvement are discussed in subsequent Section~\ref{sec:discussion}.
\begin{figure*}[t!]
    \centering
    \includegraphics[width=0.98\textwidth]{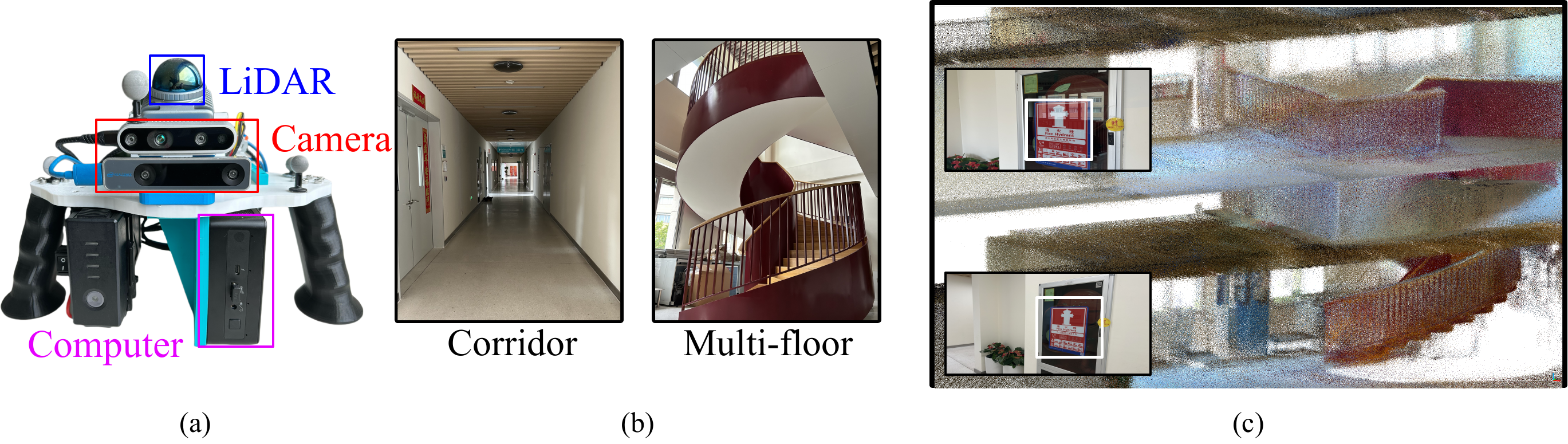}
    \caption{\textbf{Handheld Device and Self-collected Dataset.} (a) Our handheld device for collecting the Cross-floor dataset. (b) Snapshots of the Cross-floor dataset. (c) Snapshots of the 3D structure of
    the Cross-floor dataset. Different levels of the environments depict a similar
    appearance and 3D geometry. An example of a repetitive scene on different levels of floors is shown. The white bounding box indicates that the repeated text appears at different locations.}
    \label{fig:handheld}
\end{figure*}
\begin{table*}[htbp] 
\centering
\caption{\textbf{Experiment results of loop closure verification.} \textbf{Best} and \underline{second best} results are highlighted. The last column presents the average performance of all datasets. \textbf{MR} represents maximum recall @100 precision, \textbf{AP} stands for average precision. The results are rounded to two decimal places.}
\label{tab:all}
\normalsize
\small
\begin{tabular}{ccccccccccccccc}
\toprule
\multirow{2}{*}{} & \multirow{2}{*}{\textbf{Method}} & \multicolumn{2}{c}{\textbf{Hotel \cite{li2024resolving}}} & \multicolumn{2}{c}{\textbf{Warehouse \cite{park2024benchmark}}}    & \multicolumn{2}{c}{\textbf{Escalator \cite{jiao2022fusionportable}}} & \multicolumn{2}{c}{\textbf{Cross-floor}} & \multicolumn{2}{c}{\textbf{WildScenes\cite{wildscenes2024}}} &\multicolumn{2}{c}{\textbf{Average}}  \\
\multicolumn{1}{c}{} & \multicolumn{1}{c}{} & \multicolumn{1}{c}{\textbf{AP}} & \textbf{MR} & \multicolumn{1}{c}{\textbf{AP}} &\textbf{MR} &  \multicolumn{1}{c}{\textbf{AP}} & \textbf{MR} &  \multicolumn{1}{c}{\textbf{AP}} & \textbf{MR} &  \multicolumn{1}{c}{\textbf{AP}} & \textbf{MR} &  \multicolumn{1}{c}{\textbf{AP}} & \textbf{MR} \\
\midrule
& DBoW \cite{angeli2008fast} & 99.18 & 37.37 & 88.72 & \underline{76.92} & 91.03 & 5.38 & 97.08 & 18.43 & 35.49 & 19.89 & 82.30 & 39.50\\
& NetVLAD \cite{arandjelovic2016netvlad}& 89.37 & 0.29 & 45.56 & 15.38 & 82.46 & 0.0 & 80.71 & 0.0 & 2.54 & 0.0 & 60.13 & 3.92 \\
& MixVPR \cite{ali2023mixvpr} & 90.57 & 0.44 & 51.35 & 23.08 & 79.15 & 0.0 & 78.56 & 0.0 & 2.16 & 0.0 & 60.36 & 5.88 \\
& AnyLoc \cite{keetha2023anyloc} & 90.49 & 0.0 & 44.77 & 15.38 & 80.58 & 0.0 & 75.53 & 0.0 & 2.31 & 0.0 & 58.74 & 3.85 \\
& SALAD \cite{izquierdo2024optimal} & 91.74 & 0.0 & 51.72 & 23.08 & 77.83 & 0.0 & 78.82 & 0.0 & 2.15 & 0.0 & 60.45 & 5.77 \\
\multirow{-6}{*}{\rotatebox{90}{Retrieval}} & TextInPlace \cite{tao2025textinplace} & 95.23 & 0.0 & 44.83 & 0.0 & 89.04 & 0.0 & 86.30 & 0.16 & 4.17 & 0.0 & 63.91 & 0.04 \\
\midrule
& ORB-NN \cite{murORB2}& \underline{99.55} & \underline{61.38} & \underline{90.18} & \underline{76.92} & 95.55 & 32.31 & 96.61 & 36.85 & 72.46 & 26.26 & 90.87 & 58.43\\
& SIFT-NN \cite{lowe1999object} & 98.88 & 36.12 & 87.32 & 76.92 & 94.08 & 26.92 & 96.41 & 17.64 & 68.88 & 39.26 & 89.11 & 49.22  \\
& SIFT-LG \cite{lindenberger2023lightglue} & 98.85 & 13.44 & 87.67 & 69.23 & 98.03 & \textbf{63.85} & 96.88 & 46.61 & 92.78 & 49.34 & 94.84 & 60.62 \\
 & SP-SG \cite{sarlin2020superglue}& 95.24 & 2.58 & \underline{90.18} & \underline{76.92} & 94.34 & 37.69 & 97.01 & 17.95 & 77.53 & 10.61 & 90.86 & 36.44\\
& LoFTR \cite{sun2021loftr} &  98.79 & 48.24 & 86.80& \underline{76.92} & \underline{98.57} & \underline{62.31} & \underline{98.39} & 45.83 & 94.0 & 64.72 & \underline{94.85} & 66.59\\
& eLoFTR  \cite{wang2024eloftr} & 98.68 & 44.71 & 86.80 & \underline{76.92} & \textbf{98.74} & 59.23 & 96.00 & 26.46 & 93.88 & \underline{72.15} & 94.82 & \underline{69.87} \\
& DISK-LG  \cite{tyszkiewicz2020disk} & 99.43 & 33.33 & 86.63 & 61.54 & 97.17 & 53.85 & 98.23 & \underline{54.80} & 87.66 & 34.48 & 93.83 & 59.50 \\
& ALIKED-LG \cite{Zhao2023ALIKED} & 97.59 & 25.99 & 82.19 & 46.15 & 90.90 & 20.00 & 90.51 & 1.26 & 94.63& 55.17 & 91.17 & 37.14 \\
& DUSt3R \cite{dust3r_cvpr24} & 96.67 & 2.79 & 58.95 & 15.38 & 97.18 & 53.85 & 85.51 & 3.15 & 20.56 & 2.65 & 71.78 & 19.46 \\
\multirow{-10}{*}{\rotatebox{90}{Geometric Verification}} & MASt3R \cite{leroy2024grounding}& 97.24 & 10.06 & 76.82 & 38.46 & 95.29 & 4.62 & 89.05 & 6.14 & \underline{96.08} & 61.54 & 90.90 & 30.20\\ 
\midrule
& ROVER (Ours) & \textbf{99.68} & \textbf{78.12} & \textbf{100.00} & \textbf{100.00} & 97.77 & 61.54 & \textbf{99.23} & \textbf{88.03} & \textbf{99.55} & \textbf{97.88} & \textbf{99.25} & \textbf{85.11}\\
\bottomrule
\end{tabular}
\end{table*}
\subsection{Proof-of-Concept Experiments}
\label{sec:poc}
Since ROVER is only concerned with the robot's trajectory, we generate multiple
trajectory patterns to imitate the robot's operation in three common repetitive
scenes, i.e., Room, Sports Ground, and Warehouse, as shown in Fig.~\ref{fig:poc}.
We randomly generate TP and FP loop candidates, 100 each, as verification input.
Ground truth trajectories are shown in the upper left of each sub-figure of Fig.~\ref{fig:poc}, and different levels of Gaussian noise are applied to the ground truth
trajectory to simulate the odometry drift that occurs in reality. The upper
right of each sub-figure illustrates the trajectory under a Gaussian noise level of
$\sigma = 0.1$, which suggests significant drifts in odometry estimation. Under
such drift, our method still preserves competitive performance as illustrated in
the Precision-Recall Curve at the bottom. The simulated experiments demonstrate
that our proposed method is effective under the drift (up to the noise level $\sigma = 0.15$). Each of the three trajectories presents a unique motion pattern,
demonstrating that ROVER's generalizability extends to different motion patterns
of robot navigation, rather than being designed to overfit to a specific environment.
As shown in Fig.~\ref{fig:viz}, the robot's trajectory is of a highly complex
shape, which also demonstrates the effectiveness of our method in real robot navigation.
\begin{figure}[t!]
\centering
\includegraphics[width=0.48\textwidth]{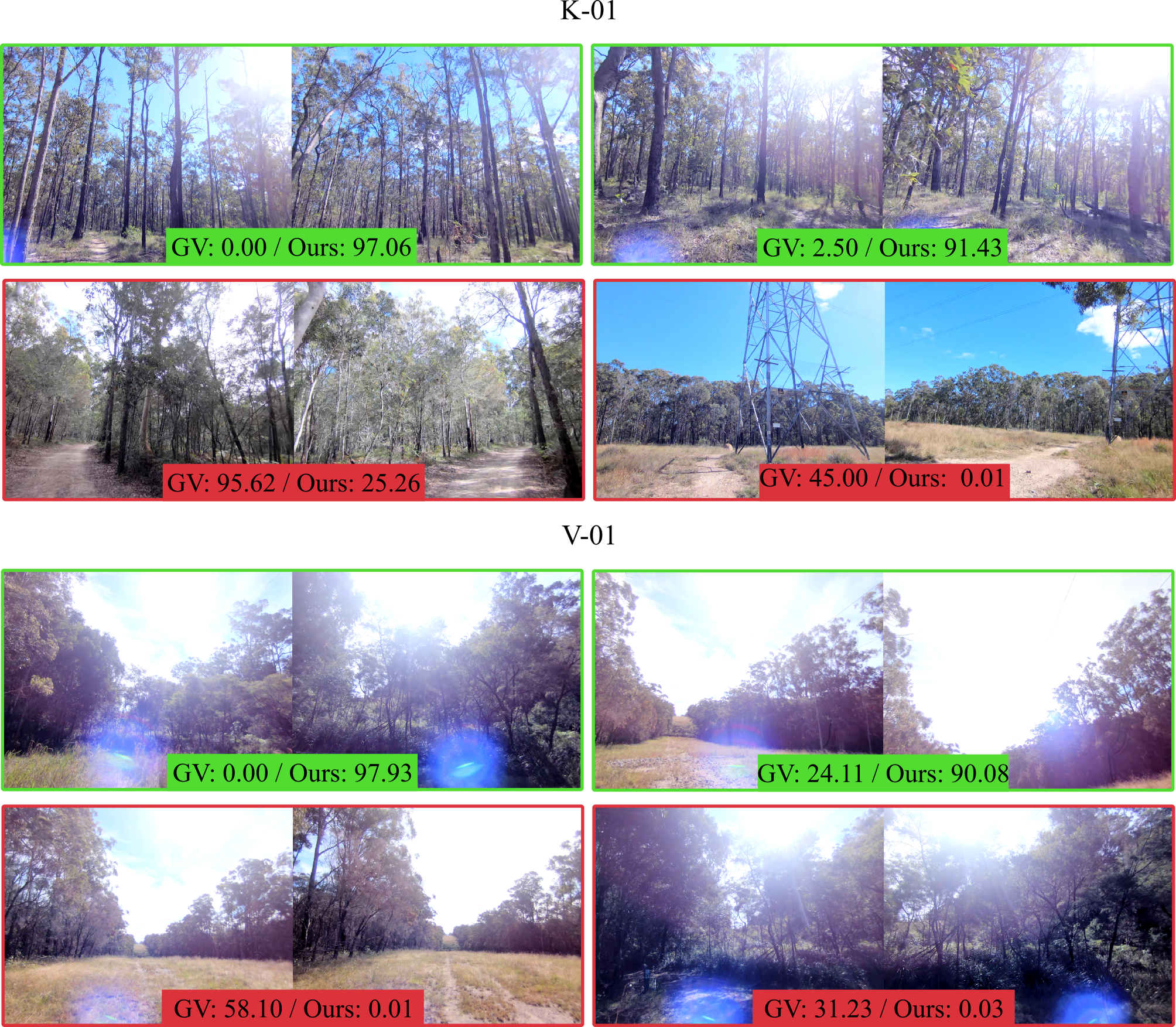}
\caption{\textbf{Pairwise loop closure verification} of K-01 and V-01 sequences on WildScenes~\cite{wildscenes2024} dataset. According to the Table~\ref{tab:all}, eloftr~\cite{wang2024eloftr} is chosen to implement geometric verification (GV). The green and red represent a true positive (\textcolor{green}{same place}) and a false positive (\textcolor{red}{different place}), respectively. In the wild scenes, the environment's appearance becomes indistinguishable from the surrounding landscape. GV fails to identify different places.} 
\vspace{-0.3cm}
\label{fig:pairwise}
\end{figure}
\subsection{Datasets and Metrics}
\label{exp:datasets} 
We evaluate our method on four public datasets: Hotel~\cite{li2024resolving},
Warehouse~\cite{park2024benchmark}, Escalator~\cite{jiao2022fusionportable},
WildScenes~\cite{wildscenes2024} and a self-collected dataset: Cross-floor.
Hotel and Warehouse datasets are collected in simulated environments. The Escalator
and WildScenes datasets are collected using handheld devices in large-scale indoor
and outdoor settings. The Cross-floor is collected in an office building with a customized handheld device. 
A quantitative evaluation of the repetitiveness of the self-collected dataset is presented in Table~\ref{tab:fpr_tpr}, where we demonstrate the FPR and TPR by varying the 
geometric verification threshold $\tau_{GV}$ with a advaned local feature matcher~\cite{edstedt2024roma}. We can observe the required False Positive Rate (FPR) needed for a functional True Positive Rate (TPR), \ie $FPR_{@TPR}$ on both datasets. For instance, to recover roughly half of the valid loops ($\text{TPR} \approx 50\%$), the Cross-floor dataset incurs a severe FPR of $36.63\%$. In contrast, the KITTI dataset achieves an even higher TPR of $66.49\%$ while maintaining a much lower FPR of $15.96\%$. This stark difference quantitatively evaluates the repetitiveness of the environment.

\begin{figure}[t]
\centering
\includegraphics[width=0.48\textwidth]{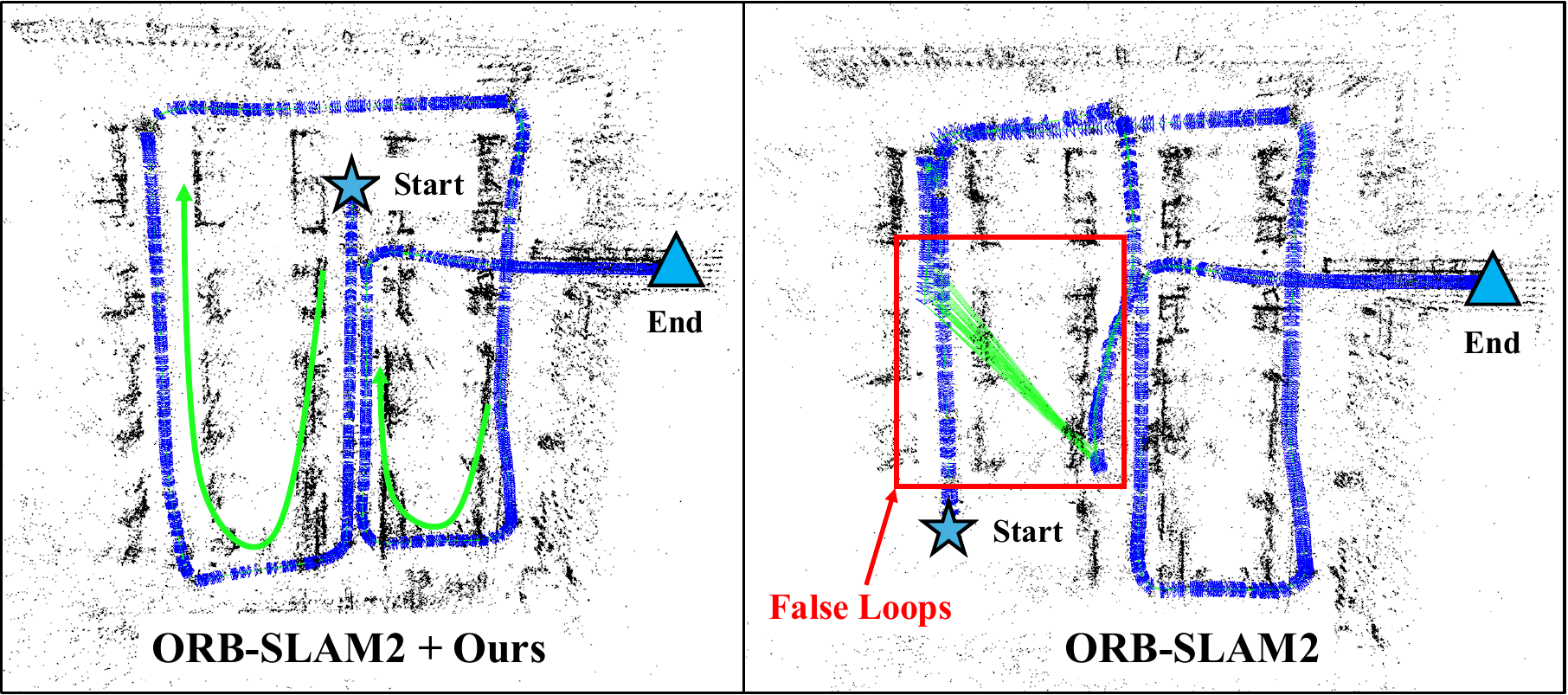}
\caption{\textbf{ORB-SLAM2's mapping result} with (left) and without (right) the ROVER of the Warehouse dataset. There are no revisits in the dataset. However, due to the high visual similarity between storage racks, false loops occur. Thus, ORB-SLAM2, combined with the ROVER method, produces a more accurate reconstruction.}
\label{fig:map}
\end{figure}

\begin{itemize}
    \item \textbf{Hotel dataset} comprises a monocular stream of two traverses through six
hotel rooms on the same floor, all with similar appearances.
    \item \textbf{Warehouse dataset} uses a logistic robot to collect the data, which does
not contain revisits (i.e., true positive loops) as shown in Fig.~\ref{fig:map}.
    \item \textbf{Escalator dataset} was collected in the real world using a handheld device that traversed four levels of an indoor building for two rounds. As shown in Fig.~\ref{fig:viz}(d), scenes with no text and repeated objects, which exist in the Escalator dataset, pose significant challenges for text- and semantics-inspired methods~\cite{li2024resolving, jin2024robust}.
    \item \textbf{WildScenes dataset} contains 5 traverses in a forest with revisits, the challenging repetitive appearance of vegetations poses a significant challenge for GV as shown in Fig.~\ref{fig:pairwise}.
    \item \textbf{Cross-floor dataset} captures a typical repetitive scene characterized by long corridors and cross-floor traverses, where the layouts of different floors are highly similar. Notably, it can be challenging for humans to distinguish between the corridors without relying on text signs. However, as shown in the Fig.~\ref{fig:handheld}(c), repeated text in the environment introduces difficulties for methods that heavily depend on scene text. To validate this observation, TextInPlace~\cite{tao2025textinplace}, which leverages scene text for reranking, is evaluated alongside appearance-only baselines.
\end{itemize}

\subsubsection{Customized Hardware}
\label{sec:hardware}
As illustrated in Fig.~\ref{fig:handheld}, we constructed a customized hardware platform for data collection featuring a Livox Mid360 3D LiDAR, an Intel RealSense D435i (RGB-D), and an Intel RealSense T265 (stereo fisheye), with a baseline of 64±0.15 mm. For the self-collected dataset, we use the visual-inertial data from T265 for its sufficient FOV in indoor low-texture environment.
Data recording was performed on an onboard PC equipped with an Intel i7 processor, 64GB of DDR4 memory, and a 1TB SSD, running Ubuntu and the Robot Operating System (ROS) for sensor synchronization. To generate the ground truth trajectory for evaluation and annotation of loop candidates, we utilized Fast-LIO2~\cite{xu2022fast}. For the localization accuracy evaluation in Table~\ref{tab:ate}, we run VINS-Fusion using the stereo fisheye stream, which is chosen to mitigate tracking failures in textureless environments where the limited field of view (FOV) of the pinhole camera (D435i) proved insufficient.

\subsubsection{Metrics}
We evaluate the proposed method using task-specific metrics. For loop closure verification, the objective is to robustly determine true positives (TP) over all the retrieved loop candidates. We follow GV-Bench~\cite{yu2024gv} in using maximum recall at 100 precision (MR) and average precision (AP) as evaluation metrics. MR represents the maximum percentage of true loops that can be identified while rejecting all the false samples. For localization accuracy, the commonly used root-mean-square error (RMSE) of absolute trajectory error (ATE) is adopted.

\begin{table}[ht]
    \centering
    \normalsize
    \caption{Comparison of False Positive Rate (FPR) and True Positive Rate (TPR) across varying GV thresholds on the self-collected Cross-floor and KITTI datasets.}
    \begin{tabular}{ccccc}
    \toprule
        \multirow{2}{*}{$\tau_{GV}$} & \multicolumn{2}{c}{\textbf{Cross-floor}} & \multicolumn{2}{c}{\textbf{KITTI~\cite{geiger2012we}}} \\
        \cmidrule(lr){2-3} \cmidrule(lr){4-5}
        & FPR & TPR & FPR & TPR \\
    \midrule
        1000 & 100.00\% & 99.21\%  & 33.66\%  & 79.10\% \\
        1200 & 98.02\%  & 93.86\%  & 15.96\%  & 66.49\% \\
        1600 & 36.63\%  & 51.65\%  & 4.45\%   & 12.95\% \\
    \bottomrule
    \end{tabular}
    \label{tab:fpr_tpr}
\end{table}
\subsection{Experiments on Loop Closure Verification}
\label{exp:LCV}
We compare our method with two types of appearance-based baselines: \textbf{global
retrieval} and \textbf{geometric verification} using local features.
\subsubsection{Global Retrieval}
We use the Euclidean distance
$d(\mathbf{v_q}, \mathbf{v_r})$ as the verification score, where $\mathbf{v_q}, \mathbf{v_r}$
represent the global descriptors of query and database images, respectively. All
the baselines, including the text-based two-stage VPR method TextInPlace~\cite{tao2025textinplace},
demonstrate significant limitations under repetitive environments. It is also
confirmed that the verification stage is a must for the LCD. 
\subsubsection{Geometric Verification}
\begin{figure*}[ht]
\centering
\includegraphics[width=0.98\textwidth]{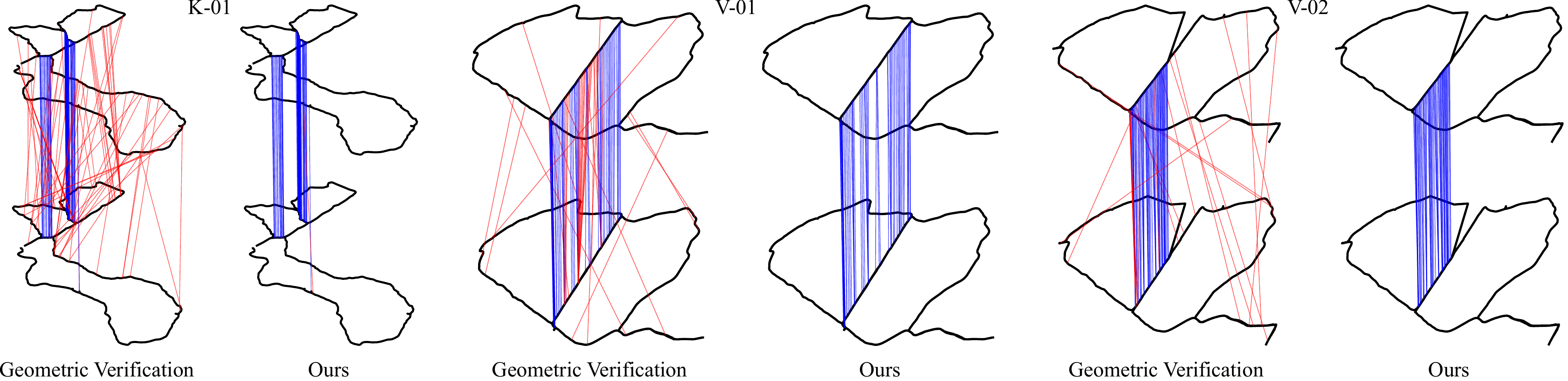}
\caption{\textbf{Visualization of loop verification result on WildScenes~\cite{wildscenes2024} dataset.} The \textcolor{blue}{true positive (TP)} and \textcolor{red}{false positive (FP)} loop closures are highlighted. The loop candidates are retrieved by SALAD~\cite{izquierdo2024optimal} and verified using geometric verification (implemented using eloftr~\cite{wang2024eloftr}) and our method. Our method consistently achieves superior performance over GV.}
\label{fig:wildscenes}
\end{figure*}

The number of RANSAC-filtered inlier matches is
used as the verification score. We compare our approach with different baselines
of local feature matching, including 3D geometric foundation models: DUSt3R~\cite{dust3r_cvpr24}
and MASt3R~\cite{leroy2024grounding}. ROVER achieves robust performance across
all datasets with the highest average AP and MR. For the Escalator dataset,
although the VINS-Fusion's odometry estimation is noisy and unstable, ROVER still
achieves competitive performance compared to costly feature matchers SIFT-LightGlue~\cite{lindenberger2023lightglue},
LoFTR~\cite{sun2021loftr}, and eLoFTR~\cite{wang2024eloftr}. More visualization
of loop verification results is shown in Fig~\ref{fig:wildscenes}.
\begin{figure*}[ht]
\centering
\includegraphics[width=0.98\textwidth]{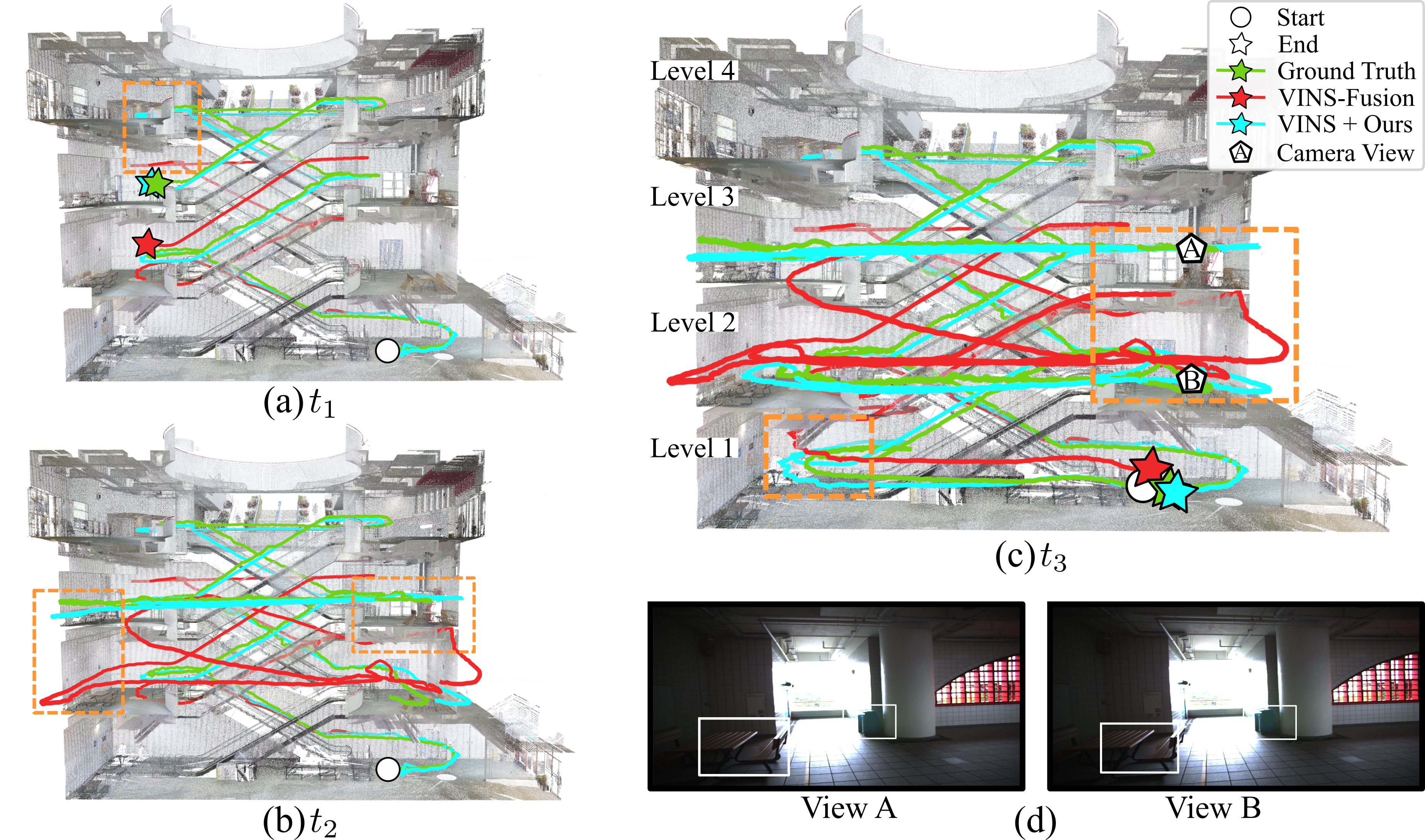}
\caption{\textbf{Trajectory comparison of the Escalator dataset.} The colored point cloud is used for visualization only. The orange dashed boxes highlight the part where real-time localization safety is critical, where the VINS-Fusion's estimated trajectory largely differs from the ground truth trajectory over different levels. VINS-Fusion, combined with our method, produces reliable real-time localization for safe robot navigation. (a)-(c) demonstrates the robot trajectory at different time $t_{i}$. (d) is an example pair of the repetitive scene in the Escalator dataset. The views A and B were captured as illustrated in (c), at different locations. The while bounding boxes highlight repeated objects over different locations.}
\label{fig:viz}
\end{figure*}
\begin{table}[htbp] 
{\centering
\caption{The comparisons of the RMSE (meters) of ATE of SLAM systems with and without the proposed method. \textbf{Best} results are highlighted. \XSolidBrush indicates tracking failure, or the datasets only provide limited sensor modality for VINS-Fusion.}
\label{tab:ate}
\scriptsize
\begin{tabular}{cccccc}
\toprule
\multicolumn{2}{c}{Method} &  Hotel & Warehouse & Escalator & Cross-floor\\
\midrule
&ORB-SLAM2& 0.67 & 0.56 & \XSolidBrush & \XSolidBrush  \\
\multirow{-2}{*}{\rotatebox{90}{VO}}  &VINS-Fusion & \XSolidBrush & \XSolidBrush & 1.14 & 0.48  \\
\midrule
 & ORB-SLAM2 & 2.17 & 15.50 & \XSolidBrush & \XSolidBrush \\
 & VINS-Fusion & \XSolidBrush & \XSolidBrush & 2.49 & 1.99 \\
\cmidrule(l){2-6}
&ORB-SLAM2 + Ours& \textbf{0.11} & \textbf{0.56} & \XSolidBrush & \XSolidBrush \\
\multirow{-4}{*}{\rotatebox{90}{SLAM}}&VINS-Fusion + Ours& \XSolidBrush & \XSolidBrush & \textbf{1.00} & \textbf{0.40} \\
\bottomrule
\end{tabular}
\\
}
\vspace{-0.5cm}
\end{table}
\subsection{Experiments on Localization Accuracy}
\label{exp:SLAM} 
\begin{figure}[h]
\centering
\includegraphics[width=0.45\textwidth]{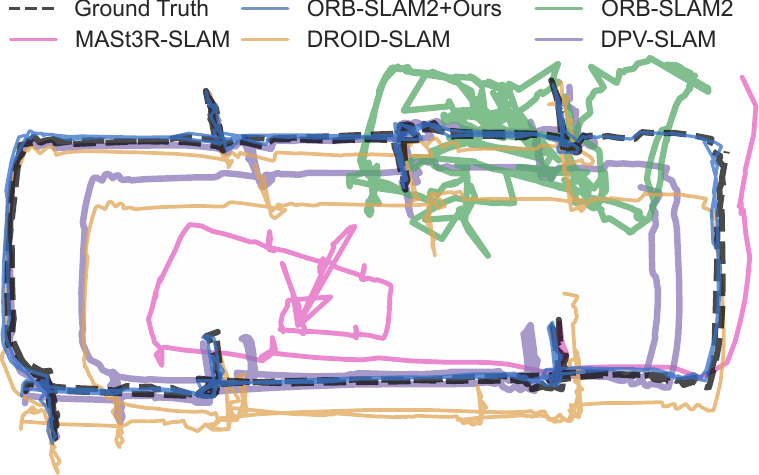}
\caption{\textbf{Trajectory comparison of the Hotel dataset.} Existing SotA visual SLAMs face challenges under repetitive scenes. Our method, combined with ORB-SLAM2, achieves the best performance thanks to robust and accurate FP loop verification.}
\label{fig:trajectory}
\end{figure}

We integrate ROVER with ORB-SLAM2 and VINS-Fusion to evaluate the
localization accuracy as shown in Table~\ref{tab:ate}. We compare the ATE over
two settings: i) Visual Odometry (VO) and ii) SLAM. For ORB-SLAM2, VO denotes a tailored
ORB-SLAM2 with \textit{LoopClosing} thread disabled. For VINS-Fusion, \textit{LoopFusion}
thread is optional in the original implementation. We can see that the SLAM
systems without ROVER, overall, exhibit huge errors due to false positive (FP)
loops caused by repetitive patterns in the scenes. However, ROVER is capable of
rejecting those FPs to get a more accurate localization performance. In the Warehouse
dataset, due to the absence of revisits, ORB-SLAM2 with ROVER achieves the identical
result as VO. We also compare our approach with various Mono-SLAM algorithms, including
learning-based methods, as shown in Fig.~\ref{fig:trajectory}. The results
demonstrate that repetitive scenes pose substantial challenges for learning-based
algorithms as well. Furthermore, we compare our method with offline robust
optimization backends in Fig.~\ref{fig:backends}, ROVER enables robust
localization performance in real-time SLAM.
\subsubsection{Localization safety}
We evaluate the real-time localization safety by
measuring ATE once in a fixed time interval in Table~\ref{tab:tate}. We conducted
the experiment over the Escalator dataset, where the whole traverse is divided
into 5 evenly distributed intervals. At $t_{1}$, $t_{2}$, and $t_{3}$, the VINS-Fusion
shows significantly worse ATE compared to the ATE of the entire traverse at
$t_{4}$. For the three timestamps shown in the Fig.~\ref{fig:viz}, VINS-Fusion estimates
false localization across different floors, which can be fatal for robotic
operations, depicting the deficiency of ATE evaluation over the complete traverse.
Therefore, we propose to compute a temporal ATE (tATE):
\begin{equation}
    tATE = \sqrt{1/k \sum e(\mathcal{T}_{t_i}- \mathcal{T}^{gt}_{t_i})^{2}}, \ t_{i}
    \in \{t_{0}, ..., t_{k-1}\},
\end{equation}
aiming to reflect the real-time localization safety.
\begin{table}[htbp] 
{\centering
\caption{The comparisons of the RMSE (meters) of tATE of VINS-Fusion w/ and w/o the proposed method over the Escalator dataset. \textbf{Best} results are highlighted.}
\label{tab:tate}
\begin{tabular}{cccccccc}
\toprule
Method &  $t_{0}$ & $t_{1}$ & $t_{2}$ & $t_{3}$ & $t_{4}$ & tATE \\
\midrule
VINS-Fusion (VO) & 0.70 & 0.94  & 1.22 &  1.88 & 1.14 & 1.19 \\
VINS-Fusion  & 0.70 & 4.73 &  3.89 & 2.90 & 2.49 &  3.05\\
VINS-Fusion + Ours & \textbf{0.70} & \textbf{0.94} & \textbf{1.19} & \textbf{1.30} & \textbf{1.00} & \textbf{1.03}  \\
\bottomrule
\end{tabular}
\\
}
\end{table}
\subsection{Efficiency Evaluation}
\label{sec:eff}
\begin{figure}[ht]
\centering
\includegraphics[width=0.45\textwidth]{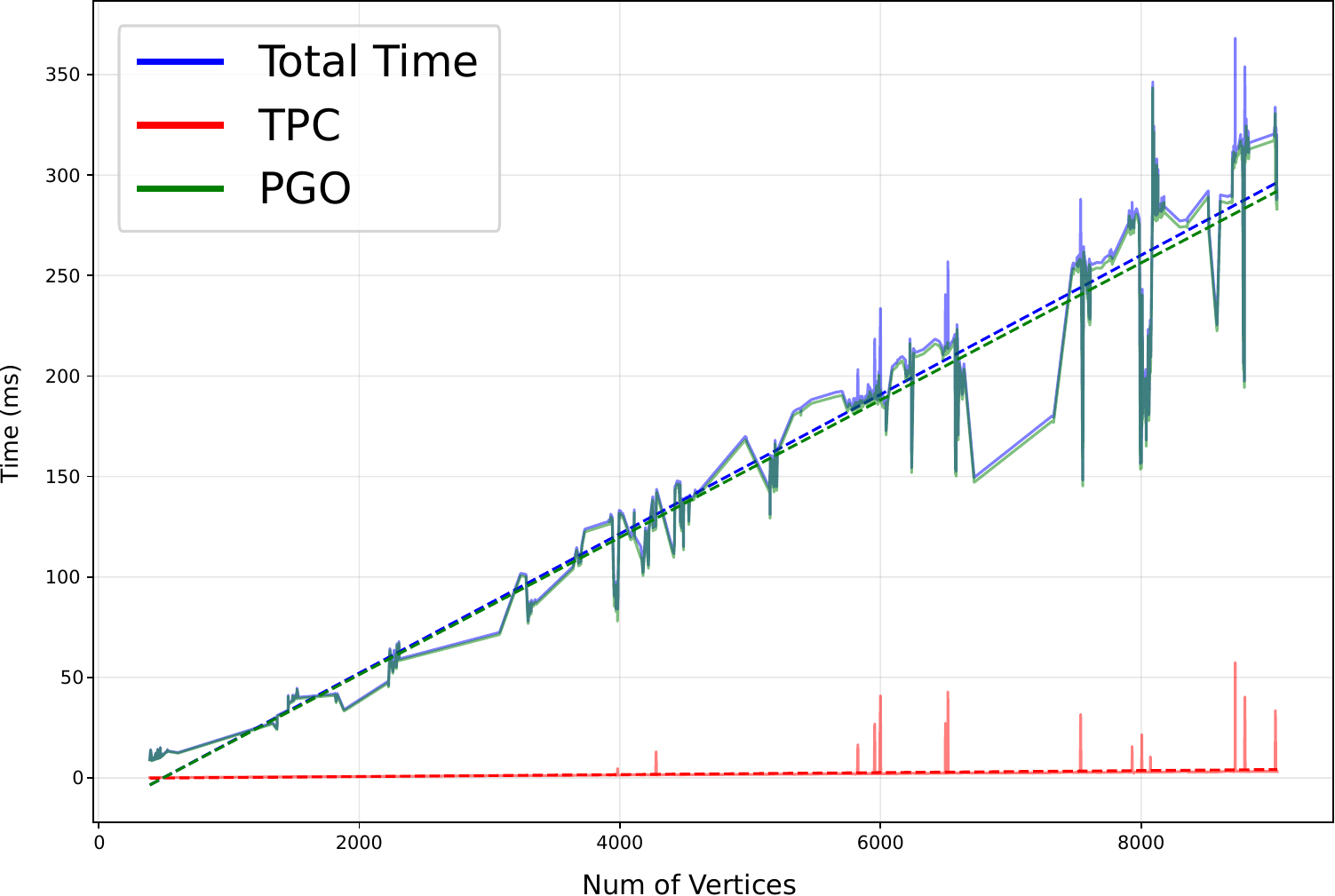}
\caption{\textbf{Runtime analysis of the algorithm.} The dashed lines represent the trend of the computation time over the number of vertices. Our proposed trajectory prior constraint (TPC) incurs only minor computational overhead in a very large environment with more than 6,000 vertices in the pose graph. Major computation time is attributed to pose graph optimization (PGO), which is initially required for loop closing in SLAM systems.} 
\label{fig:runtime}
\end{figure}

In real-world applications, both robustness and efficiency are paramount for
SLAM. We evaluate efficiency in Fig.~\ref{fig:runtime} by varying the number of
vertices from 0 to 10,000 using the $g^{2}o$~\cite{kummerle2011g} backend. A trajectory
prior constraint (TPC) is added to the pose graph for each additional vertex.
Even in large environments with over 6,000 vertices, the proposed method incurs only
minor computational overhead. Instead, the majority of the runtime is dominated by
pose graph optimization (PGO), a fundamental requirement for graph SLAM systems.
In Fig.~\ref{fig:runtime}, the red line indicates the TPC computation time defined
in Equation (\ref{equ:score}), which increases with the size of the pose graph. The
dashed lines illustrate the trends for total runtime and PGO runtime, both of
which grow linearly with the number of vertices. 

\begin{figure}[ht]
\centering
\includegraphics[width=0.48\textwidth]{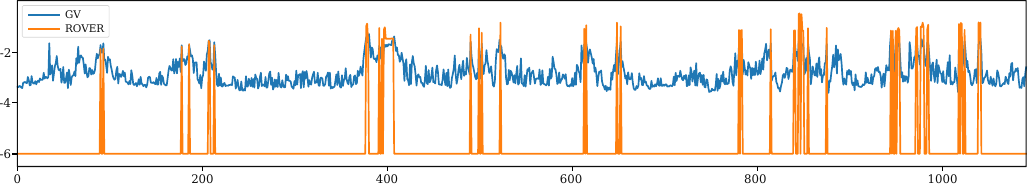}
\caption{\textbf{Time-to-Verify} per loop in the Hotel dataset. To better illustrate the relationship between geometric verification (GV) and ROVER runtimes, the x-axis represents the detected loop index, and the y-axis shows the logarithm of the raw runtime. The average runtime for ROVER and GV is 1.91 ms and 1.63 ms, respectively.}
\label{fig:loop_runtime}
\end{figure}

Furthermore, we measure the ``Time-to-Verify'' per loop to validate that integrating ROVER to the SLAM process does not break the real-time constraint for the primary threads. As shown in Fig.~\ref{fig:loop_runtime}, the runtime of ROVER is comparable to GV, while in real-world cases, ROVER is only triggered when a loop candidate passes the GV, leading ROVER to have a comparable average runtime per loop of 1.91 ms in comparison to 1.63 ms for GV.
A detailed runtime analysis of ROVER integrating with SLAM is provided in Table~\ref{tab:runtime}. As shown, integrating ROVER into both original SLAM threads results in minimal FPS drops, preserving the real-time constraint.
The run time discrepancy between ORB-SLAM2 and VINS-Fusion is attributable to their different solver backends: ORB-SLAM2 utilizes $g^{2}o$, whereas VINS-Fusion relies on Ceres~\cite{Agarwal_Ceres_Solver_2022}. As $g^{2}o$ is specifically optimized for graph optimization, it demonstrates superior performance in this context. 
In conclusion, ROVER introduces minimal cost because loop closures in real-world scenarios are sparse. In conclusion, integrating ROVER into real-time SLAM systems ensures robust loop closure detection while preserving efficiency.
\begin{table}[htbp] 
{\centering
\caption{The runtime comparison of ORB-SLAM2 and VINS-Fusion with and without the proposed method. We measure the median and mean processing times in milliseconds, as well as the average frame rate (FPS).}
\label{tab:runtime}
\begin{tabular}{cccc}
\toprule
Method &  Median (ms) $\downarrow$ & Mean (ms) $\downarrow$  &  FPS $\uparrow$ \\
\midrule
ORB-SLAM2  & 1.28 & 1.34  & 74.90\\
ORB-SLAM2 + Ours  & 1.36 \textcolor{gray}{(+6\%)} & 1.44 \textcolor{gray}{(+7\%)} & 69.23 \textcolor{gray}{(-7\%)}\\
\midrule
VINS-Fusion  & 38.95 & 42.81 & 2.34 \\
VINS-Fusion + Ours  & 43.24 \textcolor{gray}{(+11\%)} &43.64 \textcolor{gray}{(+2\%)} & 2.29 \textcolor{gray}{(-2\%)} \\
\bottomrule
\end{tabular}
\\
}
\end{table}

\begin{figure*}[h]
\centering
\includegraphics[width=0.96\textwidth]{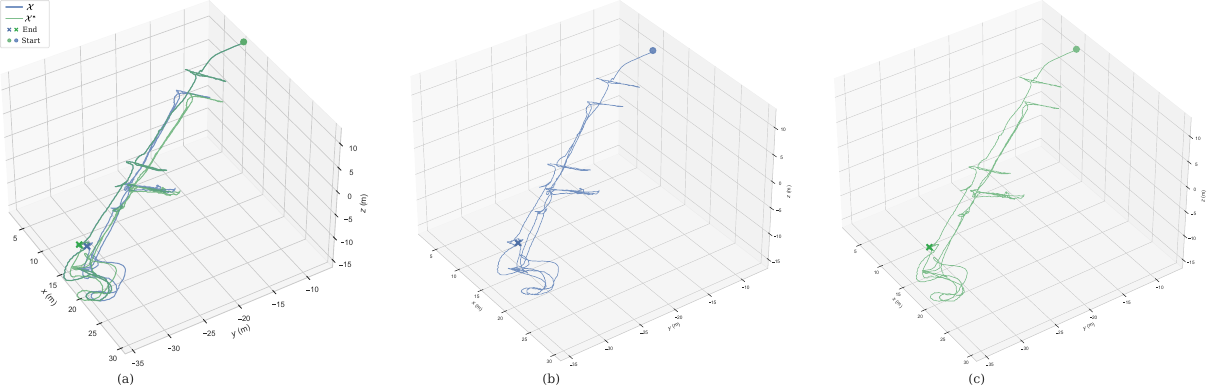}
\caption{\textbf{Visualization of a false positive loop correctly rejected by ROVER on the Cross-floor dataset.} (a) Combined plot of the two trajectories, (b) the trajectory before optimization $\mathcal{X}$, and (c) the trajectory after optimization $\mathcal{X}^*$.}
\label{fig:fn_traj}
\end{figure*}

\section{Discussion} 
\label{sec:discussion}

\subsection{Suboptimal Cases} 
\label{sec:dis_subopt}
As discussed in Fig.~\ref{fig:poc} and Section~\ref{exp:LCV}, ROVER achieves suboptimal performance due to significant drift in the front-end's trajectory estimation. Since the method relies on the historical trajectory as a prior constraint, substantial odometry errors can mislead the verification process. When integrating ROVER into SLAM systems, we apply a strict threshold to maintain a conservative approach, accepting only loop closure candidates with high confidence. 
This strategy ensures the robustness of the SLAM system by prioritizing 100\% precision, though it inevitably rejects some valid loops that could otherwise contribute to drift correction.
A false negative case is shown in Fig.~\ref{fig:fn_traj} to better illustrate this case in the Cross-floor dataset. Since we use a strict threshold in the indoor scenario, though this true loop contributes to the PGO, it is rejected.
However, as shown in Table~\ref{tab:ate}, this trade-off still yields superior results in most scenarios, particularly when compared to GV.

However, relying on a fixed threshold is suboptimal for long-term operation. As the SLAM system's running time increases, accumulated drift grows, meaning the uncertainty of the trajectory estimate varies over time. A fixed threshold acts as a static switch for trusting the trajectory prior, which fails to account for the degradation of the prior's reliability over time. Consequently, while a strict fixed threshold prevents false positives, it may become overly restrictive in later stages of navigation, limiting the potential for loop closure recall.

\subsection{Thresholding Function}
\label{sec:thresholding_function}
\begin{figure}[h]
\centering
\includegraphics[width=0.48\textwidth]{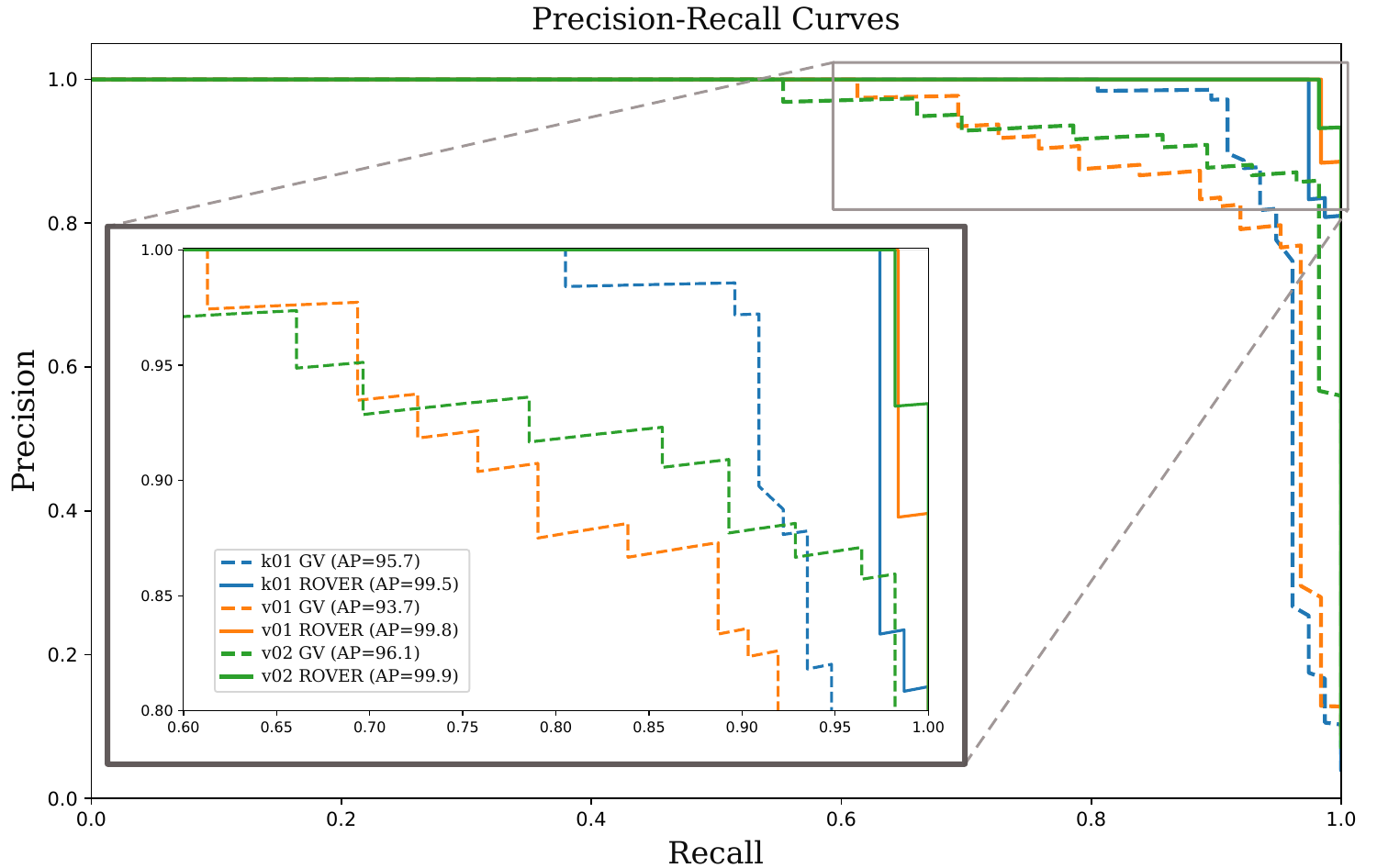}
\caption{\textbf{PR-Curves on the WildScenes Dataset.} 
Performance comparison of the proposed ROVER (solid lines) against the GV baseline (dashed lines) across the k01, v01, and v02 sequences. Each sequence is denoted by a distinct color. The Precision-Recall curves demonstrate that ROVER consistently outperforms GV across the entire threshold spectrum, highlighting its superior robustness in diverse outdoor environments.
}
\label{fig:pr-curve}
\end{figure}

Equation~\ref{equ:rover} introduces the thresholding function $\Psi$, a critical component for real-world deployment. In our experiments, we employ a truncation function with environment-specific thresholds: $\tau = 0.5$ for indoor and $\tau = 1.0$ for outdoor settings. This parameterization accounts for the front-end uncertainty, which empirically scales with the size and complexity of the environment. While SLAM front-end uncertainty estimation can be inherently noisy, the Precision-Recall (PR) curves in Fig.~\ref{fig:pr-curve} demonstrate the robustness of our method. Specifically, the curves show that ROVER (solid lines) consistently outperforms traditional Geometric Verification (GV, dashed lines) across the entire threshold spectrum. Because existing SLAM systems also rely on empirically chosen GV thresholds, demonstrating consistent improvement across the full PR curve provides strong evidence that ROVER fundamentally achieves superior binary classification in repetitive scenes, regardless of the chosen threshold. Future work will explore integrating an online uncertainty estimator to adaptively determine this optimal threshold in real time.
\subsection{System Integration}
\label{sec:dis_system}
In this section, we present a detailed ablation study that systematically varies the GV threshold, $\tau_{GV}$ in Table~\ref{tab:ablation}. The standard default value in ORB-SLAM2~\cite{murORB2} is set to $\tau_{GV}=40$, as implemented in our sequential pipeline (Algorithm~\ref{alg:loopclosing}). Interestingly, our quantitative results demonstrate that omitting the GV module entirely (\ie setting $\tau_{GV}=0$) yields the lowest ATE. This finding strongly suggests that ROVER is highly capable of recovering valid loop closures that are otherwise incorrectly rejected by GV. This advantage is particularly evident in challenging scenarios, such as low-texture environments, where feature-based GV frequently fail to find sufficient inliers, resulting in false negatives. However, completely bypassing GV and relying solely on ROVER introduces a critical trade-off that renders the overall system overly sensitive to the selection of ROVER's own threshold, as discussed in the preceding section. Consequently, a highly valuable direction for future work involves fundamentally refining the decision logic outlined in Algorithm~\ref{alg:loopclosing}. Rather than employing a sequential filtering pipeline, a more robust approach would evaluate traditional geometric consistency (GV) and our trajectory prior (ROVER) in parallel. Such a parallel fusion architecture would allow the system to safely and confidently recover false negatives without sacrificing the inherent stability provided by geometric constraints.
\begin{table}[h]
    \centering
    % \resizebox{0.38\textwidth}{!}{
    \normalsize
     \caption{\textbf{Ablation Study on $\tau_{GV}$.} We demonstrate the localization accuracy on the Hotel~\cite{li2024resolving} by varying the GV threshold in Algorithm~\ref{alg:loopclosing}.}
    \begin{tabular}{cccc}
    \toprule
         $\tau_{GV}$ & RMSE & $t_{median}$ & $t_{mean}$\\
    \midrule
         0  & \textbf{0.018385} & 1.37904 & 1.46963 \\
         10 & 0.033024 & 1.33717 & 1.42473 \\
         20 & 0.550486 & 1.34598 & 1.42309 \\
         30 & 0.684389 &\textbf{1.29997} & \textbf{1.37877} \\
         40 & 0.022741 & 1.38295 & 1.47284 \\
         50 & 0.029010 & 1.37358 & 1.45792\\
         \bottomrule
    \end{tabular}
    % }
    \label{tab:ablation}
\end{table}
\subsection{Future Work} 
\label{sec:dis_future}
As discussed above, the primary limitation of the current approach stems from the use of a fixed threshold. Loop verification is inherently a binary classification task where finding a perfect, universal scoring function is challenging. To address this, future work will focus on developing an adaptive thresholding mechanism. By incorporating an online uncertainty estimator, the system could dynamically adjust the Trajectory Prior Constraint (TPC) based on the confidence level of the state estimator. This would allow the system to relax the constraint when odometry uncertainty is high, thereby balancing the trust between the trajectory prior and visual loop candidates.
Beyond employing an adaptive threshold, future system integration should focus on a tighter fusion of GV and ROVER. Instead of the sequential approach outlined in Algorithm~\ref{alg:loopclosing}, evaluating the prior trajectory and geometric consistency in parallel would enable ROVER to effectively recover false negative loops that GV erroneously rejects.

Furthermore, we plan to investigate the integration of submap techniques, similar to those used in ORB-SLAM3~\cite{campos2021orb}. Submaps can mitigate accumulated drift by periodically resetting the local reference frame. Moreover, as shown in Fig.~\ref{fig:runtime}, the computational cost of pose-graph optimization increases linearly with the number of vertices. Adopting a submap-based verification strategy would not only mitigate drift but also bound the size of the pose graph, thereby improving real-time performance in scalable environments. In summary, our future work includes enhancing ROVER's robustness through uncertainty-aware adaptive thresholding and extending its scalability via submap-based architectures.

\section{Conclusions}
\label{sec:con} In this article, we investigate the benefit of leveraging the trajectory
prior constraint for LCV and propose ROVER, targeting repetitive environments. It
achieves robust performance compared to commonly adopted geometric verification methods
and can be seamlessly integrated with existing real-time SLAM frameworks.
The main idea is to evaluate the geometric consistency between the robot's trajectory before and after the PGO, which serves as a measure of the confidence for loop candidates.
Extensive public and self-collected benchmark results validate the effectiveness
and robustness of the proposed work. In the future, as discussed in Section~\ref{sec:discussion},
we plan to extend our framework by introducing dynamic uncertainty prediction in
odometry estimation and submap-based architectures to further improve the robustness of our LCV method in scalable repetitive environments.

\bibliographystyle{IEEEtran}
\bibliography{ref}
\vspace{-1.5cm}
\begin{IEEEbiography}[{\includegraphics[width=1in,height=1.25in,clip,keepaspectratio]{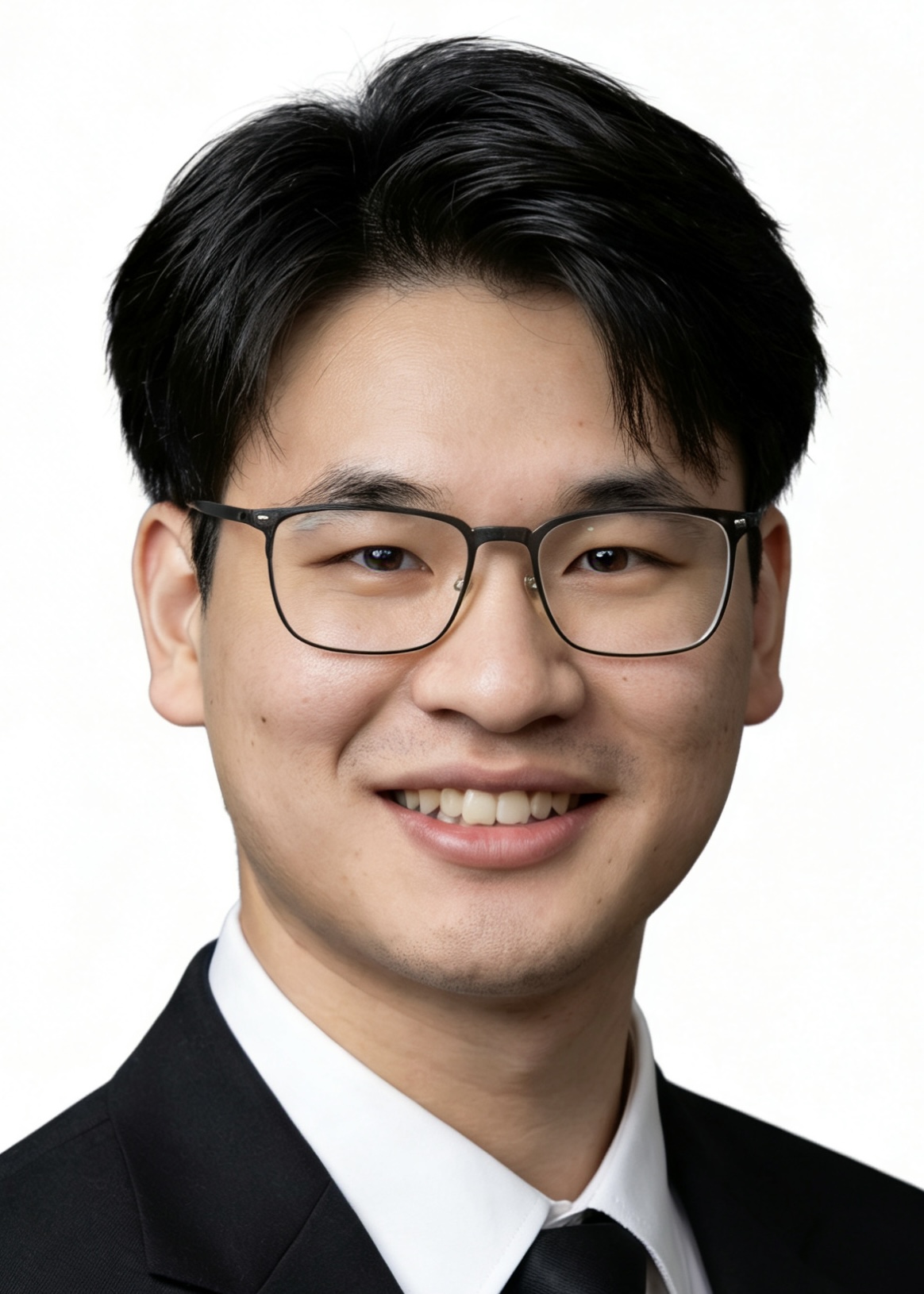}}]{Jingwen Yu} (Student Member, IEEE) received the B.E. degree in information engineering from the Southern University of Science and Technology, Shenzhen, China, in 2021. He is currently a Ph.D. candidate in the Department of Electronic and Computer Engineering at The Hong Kong University of Science and Technology, Hong Kong, China. His research interests include SLAM, mobile robot navigation, and 3D reconstruction.
\end{IEEEbiography}
\vspace{-1.5cm}
\begin{IEEEbiography}[{\includegraphics[width=1in,height=1.25in,clip,keepaspectratio]{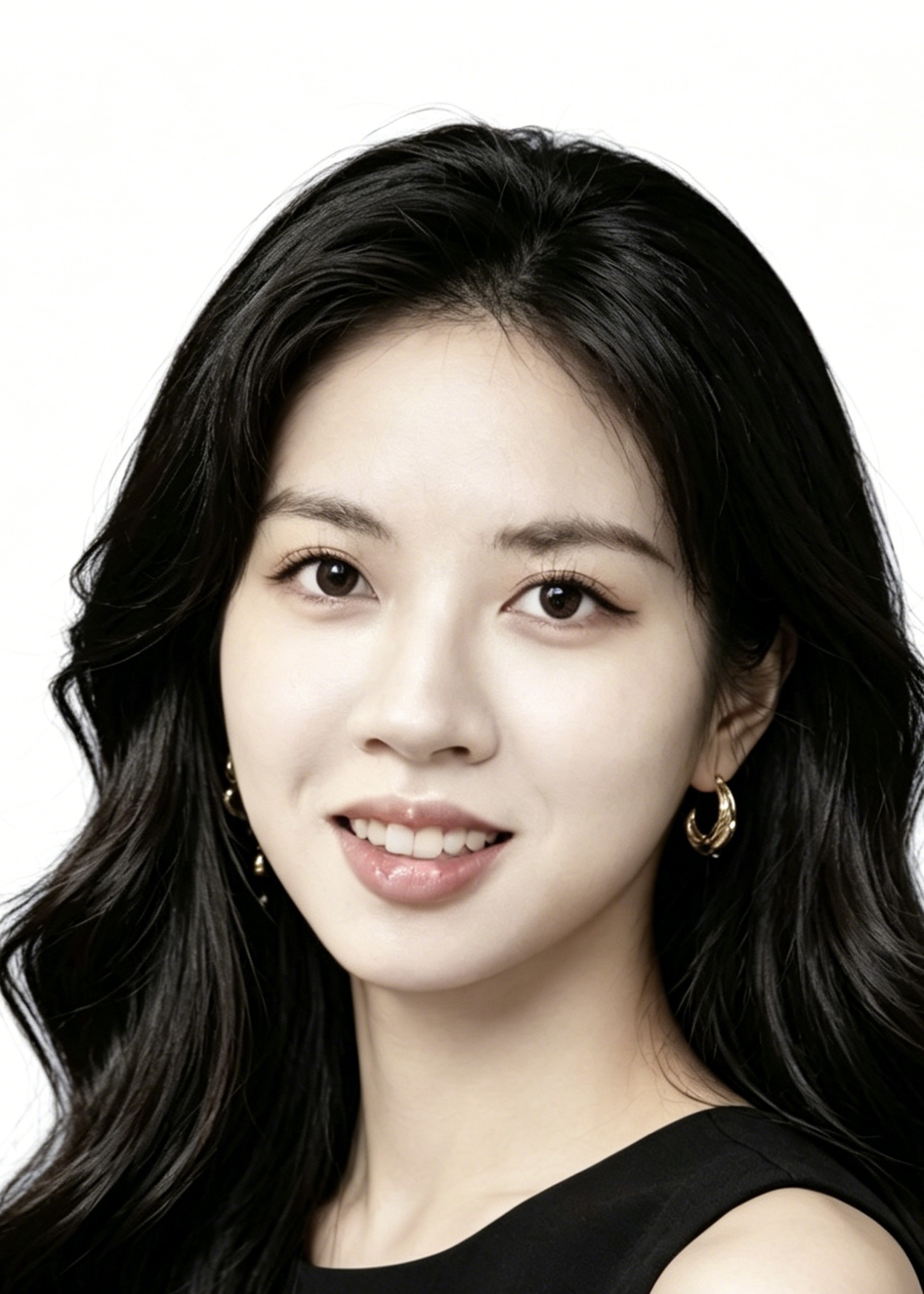}}]{Jiayi Yang} received the B.E. degree in information engineering from the Southern University of Science and Technology, Shenzhen, China, in 2024. She is currently pursuing her master's degree in the Department of Electrical Engineering and Information Systems at the University of Tokyo, Tokyo, Japan. Her research interests include simultaneous localization and mapping and human-computer interaction.
\end{IEEEbiography}
\vspace{-1.5cm}
\begin{IEEEbiography}[{\includegraphics[width=1in,height=1.25in,clip,keepaspectratio]{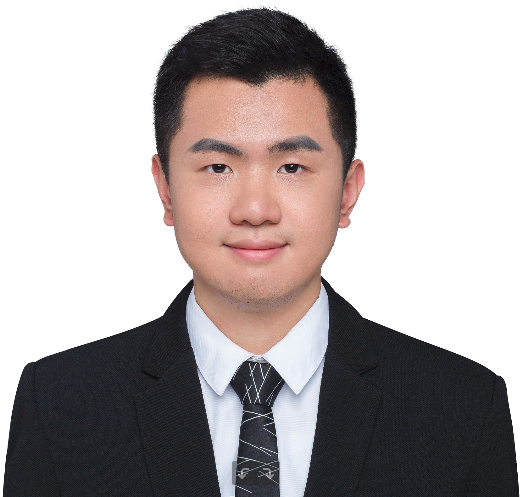}}]{Jianhao Jiao} (Member, IEEE) received the B.E. degree in instrument science from Zhejiang University, Hangzhou, China, in 2017, and the Ph.D. degree from the Hong Kong University of Science and Technology, Hong Kong, in 2021. He is currently a senior research fellow at the Department of Aeronautical and Aviation Engineering, Hong Kong Polytechnic University, Hong Kong, and an honorary senior research fellow in the Department of Computer Science at University College London, UK. His research interests include SLAM and navigation.
\end{IEEEbiography}
\vspace{-1.5cm}
\begin{IEEEbiography}[{\includegraphics[height=1.25in,clip,keepaspectratio]{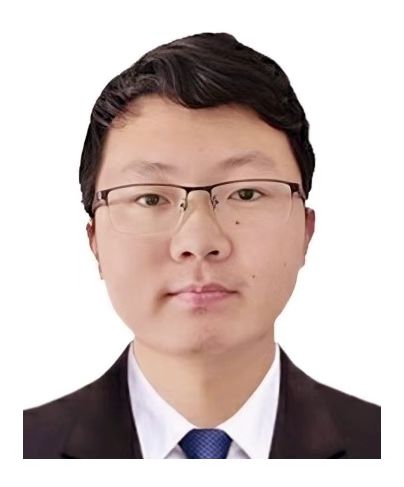}}]{Anjun Hu} received the B.E. degree in mechatronic engineering from the Department of Mechanical and Electronic Engineering, Shenyang Jianzhu University, Shenyang, China, in 2023. He is currently pursuing a Master's degree with the Department of Electronic and Electrical Engineering, Southern University of Science and Technology, Shenzhen, China. 
His research interests include robotic mechanism design and structural optimization.
\end{IEEEbiography}
\vspace{-1.5cm}
\begin{IEEEbiography}[{\includegraphics[width=1in,height=1.25in,clip,keepaspectratio]{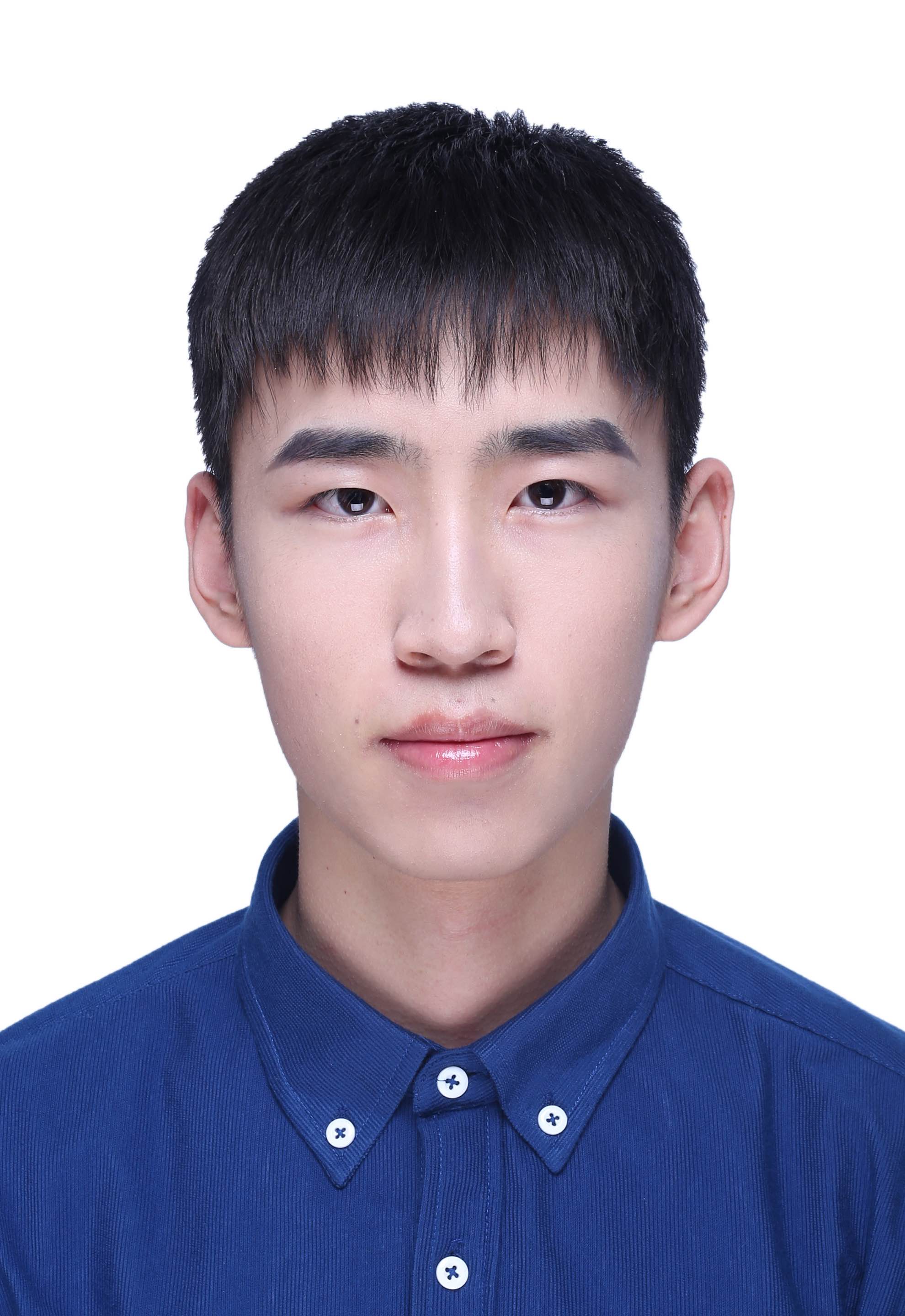}}]{Zhonghang Liu} received the B.Sc. degree from Huazhong Agricultural University, Wuhan, China, in 2020 and his Ph.D. degree from Singapore Management University, Singapore, in 2025. He is currently a researcher at SmartMore Corporation. His research interests primarily lie in computer vision and machine learning, with a focus on developing robust visual perception systems. He has published several papers in CVPR, NeurIPS, and ECCV. He also serves as a reviewer for CVPR, ICML, WACV, and TNNLS.
\end{IEEEbiography}
\vspace{-1.5cm}
\begin{IEEEbiography}
[{\includegraphics[width=1in,height=1.25in,clip,keepaspectratio]{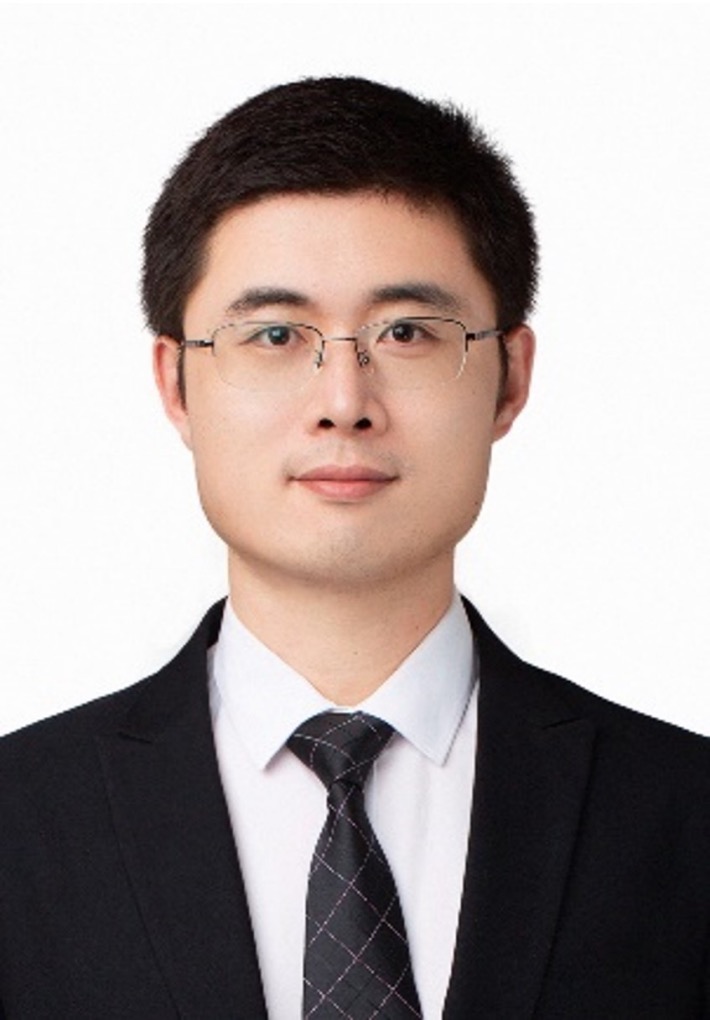}}]{Jiankun Wang} (Senior Member, IEEE) received the B.E. degree in Automation from Shandong University, Jinan, China, in 2015, and the Ph.D. degree from the Department of Electronic Engineering, The Chinese University of Hong Kong, Hong Kong, in 2019. He is currently an Assistant Professor with the Department of Electronic and Electrical Engineering, Southern University of Science and Technology, Shenzhen, China. His research interests include robot motion and path planning, human–robot interaction, and artificial intelligence. He serves as the Associate Editor of IEEE Transactions on Automation Science and Engineering, IEEE Transactions on Intelligent Vehicles, IEEE Robotics and Automation Letters, International Journal of Robotics and Automation, and Biomimetic Intelligence and Robotics.
\end{IEEEbiography}
\vspace{-1.5cm}
\begin{IEEEbiography}[{\includegraphics[width=1in,height=1.25in,clip,keepaspectratio]{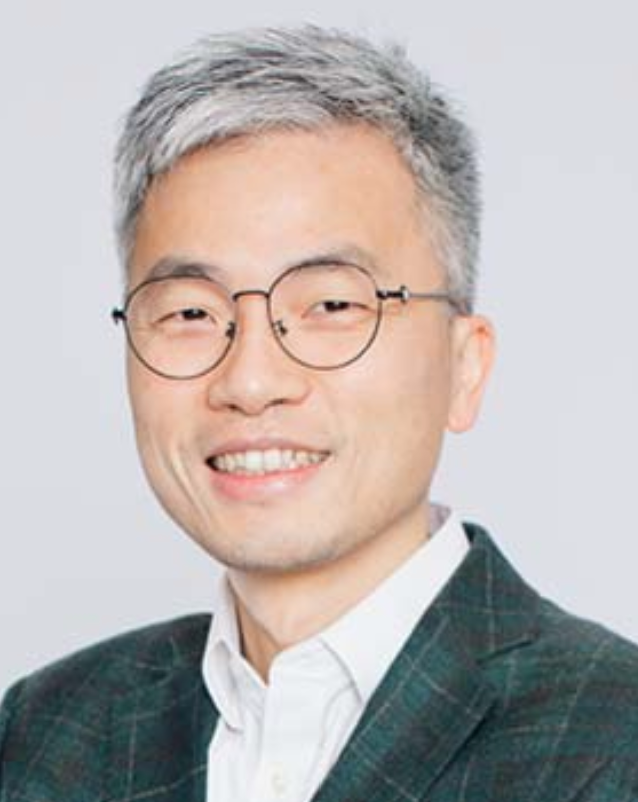}}]{Ping Tan} (Senior Member, IEEE) is a Professor with the Department of Electronic and Computer. Engineering,the Hong Kong University of Science and Technology (HKUST). Before that, he was the head of XR Lab at Alibaba DAMO Academy, an associate professor in Simon Fraser University (SFU) and the National University of Singapore (NUS). He received the master's and bachelor's degrees from Shanghai Jiao Tong University (SJTU), China, in 2003 and 2000, respectively, and the PhD degree from the Hong Kong University of Science and Technology (HKUST) in 2007.
His research interests include computer vision, computer graphics, and robotics. He was an editorial board member of the IEEE Transactions on Pattern Analysis and Machine Intelligence (TPAMI) and the International Journal of Computer Vision (IJCV).
\end{IEEEbiography}
\vspace{-1.5cm}
\begin{IEEEbiography}[{\includegraphics[width=1in,height=1.25in,clip,keepaspectratio]{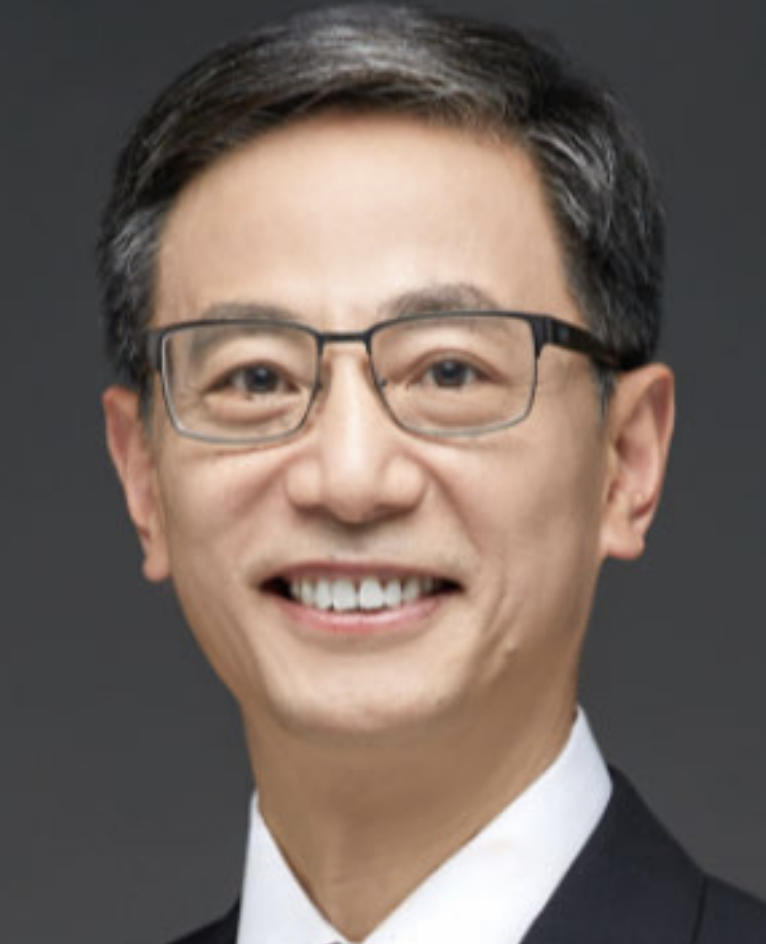}}]{Hong Zhang} (Life Fellow, IEEE) received his BSc from Northeastern University (Boston) and PhD from Purdue University, both in Electrical Engineering. From 1988-2020, he was on the faculty of the Department of Computing Science, University of Alberta, Edmonton, Canada. Dr. Zhang joined the Department of Electronic and Electrical Engineering, Southern University of Science and Technology (SUSTech) in Shenzhen, China, in 2020, where he is currently a Chair Professor and Director of the Shenzhen Key Laboratory of Robotics and Computer Vision. Dr. Zhang's research interests include robotics, computer vision, and image processing. Among his services in the international robotics community, he was the Editor-in-Chief of the IROS Conference Paper Review Board, a flagship conference of the IEEE Robotics and Automation Society (RAS), from 2020 to 2022. Currently, he is serving a two-year term as Vice President of Publications of the IEEE RAS, from 2026 to 2027. Dr. Zhang is a Fellow of IEEE and a Fellow of the Canadian Academy of Engineering.
\end{IEEEbiography}
\end{document}